\PassOptionsToPackage{table}{xcolor}
\documentclass[11pt]{article}

\usepackage[preprint]{acl}

\usepackage{times}
\usepackage{latexsym}
\usepackage[T1]{fontenc}
\usepackage[utf8]{inputenc}
\usepackage{microtype}
\usepackage{inconsolata}
\usepackage{graphicx}
\usepackage{booktabs}
\usepackage{array}
\usepackage{multirow}
\usepackage{pifont}
\usepackage{xcolor}
\usepackage{amsmath}
\usepackage{amssymb}
\usepackage{algorithm}
\usepackage{algpseudocode}
\usepackage{xspace}
\usepackage{placeins}
\usepackage{float}
\usepackage[nameinlink,noabbrev]{cleveref}
\usepackage{enumitem}

\newcommand{\benchmark}{MemTrace\xspace}
\newcommand{\fresh}{Fresh\xspace}
\newcommand{\saturated}{Saturated\xspace}
\newcommand{\deltaforget}{$\Delta$Forget\xspace}
\newcommand{\qcurrent}{Current\xspace}
\newcommand{\qhistorical}{Historical\xspace}
\newcommand{\qtrajectory}{Trajectory\xspace}
\newcommand{\yesmark}{\ding{51}}
\newcommand{\partmark}{\ensuremath{\circ}}
\newcommand{\nomark}{--}

\graphicspath{{figures/}}

\title{MemTrace: Probing What Final Accuracy Misses in Long-Term Memory}
\author{
  Xianxuan Long\textsuperscript{1} \quad 
  Zhikai Chen\textsuperscript{1} \quad 
  Shenglai Zeng\textsuperscript{1} \\
  \textbf{Shouren Wang\textsuperscript{2} \quad 
  Kai Guo\textsuperscript{1} \quad 
  Jiliang Tang\textsuperscript{1}} \\
  \textsuperscript{1}Michigan State University \quad 
  \textsuperscript{2}Case Western Reserve University \\
  \texttt{\{longxia2, chenzh85, zengshe1, guokai1, tangjili\}@msu.edu} \\
  \texttt{sxw992@case.edu}
}

\begin{document}
\maketitle

\begin{abstract}

LLM agents increasingly maintain long-term memory of user facts across sessions. Yet such memory is usually evaluated by aggregating accuracy over question rows or episodes. Because it scores question rows independently, even when several probe one fact, it cannot show how that fact behaves as conditions change. We introduce \benchmark, a benchmark whose unit of measurement is the knowledge point---a single typed fact about the user, not an individual question. \benchmark probes each fact along three controlled dimensions: memory age (how many sessions ago it appeared in the history), question type (current, earlier, or trajectory of change), and evidence condition (present, missing, or contradicted by a false premise). Evaluating 13 memory-system configurations across four paradigms, we find that similar pooled accuracy hides different failures: recovering a fact's current and earlier states does not imply tracking how it changed, and safe abstention does not imply correcting a false premise. The dominant bottleneck is evidence use, not retrieval: when systems fail, the evidence was retrievable 10$\times$ more often than it was missing---so improving memory depends on using reachable evidence, not on more storage or retrieval. 

\end{abstract}

\section{Introduction}
Large language models (LLMs) are moving from single-turn assistants to persistent agents that interact with users across many sessions~\citep{memtrack, mem2actbench, mobilemem}. In this setting, memory is not only a matter of recalling isolated facts. A useful system~\citep{hu2026memoryageaiagents, huang2026rethinkingmemorymechanismsfoundation} is expected to remember user-specific information, update it when the user changes state, and keep its answers coherent as goals and preferences evolve~\citep{memtrack,memorycd,esmemeval}. This need is now being addressed by several lines of work, including long-context models that keep more history in the input~\citep{gemini25, oai_gpt5nano, qwen3}, retrieval-augmented systems that retrieve evidence at inference time~\citep{selfrag,hipporag}, external-memory systems that maintain persistent memory stores~\citep{simplemem,memorybank}, and agentic-memory architectures that use policies or agents to manage memories across interactions~\citep{mirix,memt}.
\begin{figure}[t]
\centering
\includegraphics[width=\columnwidth]{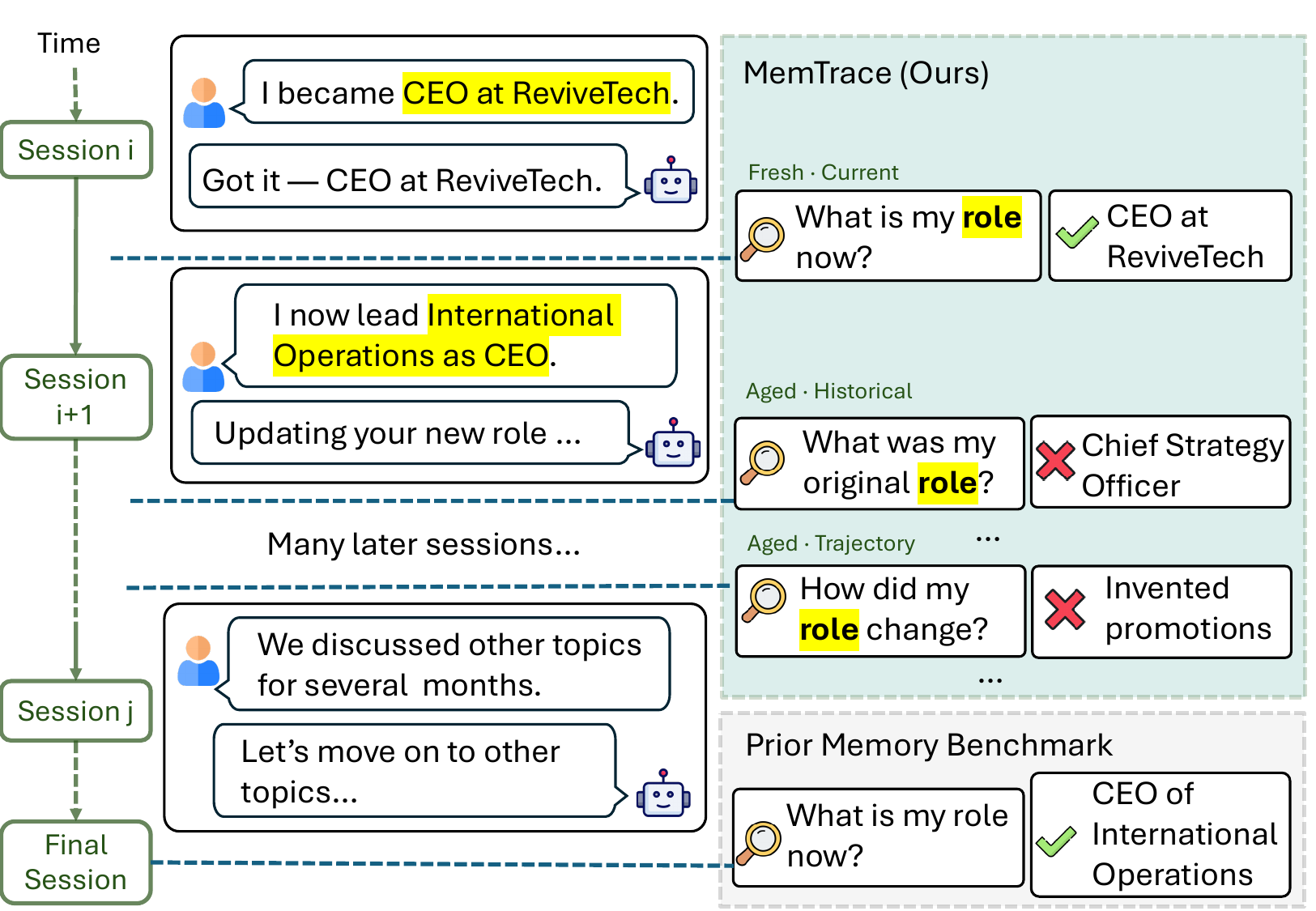}
\caption{A pooled QA view can mark one final answer as correct while hiding
failures on the same knowledge point. In this example, a system answers the
current role correctly but fails when the same role fact is probed as aged
historical and trajectory questions.}
\label{fig:teaser}
\end{figure}
Many current long-term memory benchmarks aggregate accuracy over question rows or interaction episodes~\citep{helmet,beam}, which hides variation across different queries.
This aggregation hides where a system fails, but
it is only the surface problem. The deeper one is the unit of measurement: accuracy is scored per question row or per interaction episode, so questions that probe one underlying fact are treated as independent items. As a result, no score over that unit can hold a fact fixed and ask how it behaves as conditions change. These conditions matter in practice: a system may recall a fact when recent but forget it after many sessions, answer its current state but not how it changed, or refuse an unmentioned fact yet accept a false premise about it. Capturing such contrasts requires fixing the fact and varying the conditions around it. \Cref{fig:teaser} gives an example.

Distinguishing these failures is crucial because they reflect how memory systems are used in practice. Users ask about the current state of a fact, refer back to earlier states, and ask how a fact has changed over time. They also ask questions whose evidence is absent, where the system should abstain, or whose premise conflicts with remembered information, where the system should reject the premise. Benchmarks that aggregate across these dimensions cannot show which behavior breaks, even when two systems have similar pooled scores.

We introduce \benchmark, a benchmark whose unit of measurement is the
\emph{knowledge point}: a single typed fact about the user, rather than the individual question. 
For each fact, \benchmark constructs repeated probes along controlled dimensions. \emph{Memory age} measures how many sessions have passed since the fact first appeared in history. \emph{Question type} asks for the current state, an earlier state, or the trajectory of change over time.
\emph{Evidence condition} controls whether the relevant evidence is present,
missing, or contradicted by a false premise. 
Together, these dimensions instantiate \benchmark as a benchmark of 835 typed knowledge points from 20 users, expanded into 15{,}422 question rows and over 200{,}000 scored answers.

Each user fact in \benchmark is repeatedly probed, which lets us ask whether
memory persists as sessions accumulate, whether systems track the state and evolution of a fact, and whether they behave safely when evidence is missing or conflicting. 
We also diagnose whether remaining errors come from unreachable evidence or from reachable evidence that is not used.
We evaluate 13 memory-system configurations across four paradigms on \benchmark; our key findings are as follows: \textit{(1)~Performance varies systematically across memory age, question type, and evidence condition}. Long-context systems answer recent facts well, but lose accuracy as facts age, especially on trajectory questions. RAG systems, including graph-based retrieval, handle current and earlier-state questions better than questions that require tracking change over time. Some external-memory systems decline almost all questions about facts that were never mentioned, yet rarely answer correctly when the prompt contains a false premise. \textit{(2)~Across systems, the dominant remaining gap is evidence use rather than retrieval: }when systems fail, the evidence was already retrievable about 10$\times$ more often than it was missing. Our contributions are:
\begin{itemize}[leftmargin=*,topsep=0pt,itemsep=2pt,parsep=0pt]
    \item We introduce \textbf{\benchmark}, a knowledge-point benchmark built around three probing dimensions central to long-term memory evaluation: memory age, question type, and evidence condition. These dimensions test retention, varied fact queries, and safe behavior under missing or conflicting evidence.
    \item We evaluate 13 memory-system configurations across four paradigms and show that systems with similar pooled scores fail differently. In particular, trajectory questions expose a broad weakness: systems that recover current or earlier states of a fact can still fail when asked how it changes over time.
    \item We provide diagnostic analysis of where memory failures arise. Across systems, the main bottleneck is evidence use rather than retrieval: failures come 10$\times$ more often from unused evidence than from unreachable evidence.
\end{itemize}

\vspace{-5pt}
\section{Related Work}
\label{sec:related-work}
\vspace{-2pt}
\paragraph{Memory Architectures.} Persistent LLM agents have motivated several memory architectures. Long-context models read prior interactions directly from the prompt. Retrieval-augmented systems index and retrieve external evidence before generation \citep{rag,bm25,hipporag}. Explicit or agentic memory systems maintain dedicated stores and policies for writing, updating, and retrieving memories. Some systems organize memory as an explicit agent state \citep{generativeagents,memgpt,mem0}; others study lightweight memory stores \citep{simplemem,remem,amem} or memory-management agents \citep{mirix,memt}. This architectural diversity makes final-answer accuracy an incomplete way to compare memory systems.

\begin{figure*}[t]
\centering
\includegraphics[width=0.9\textwidth]{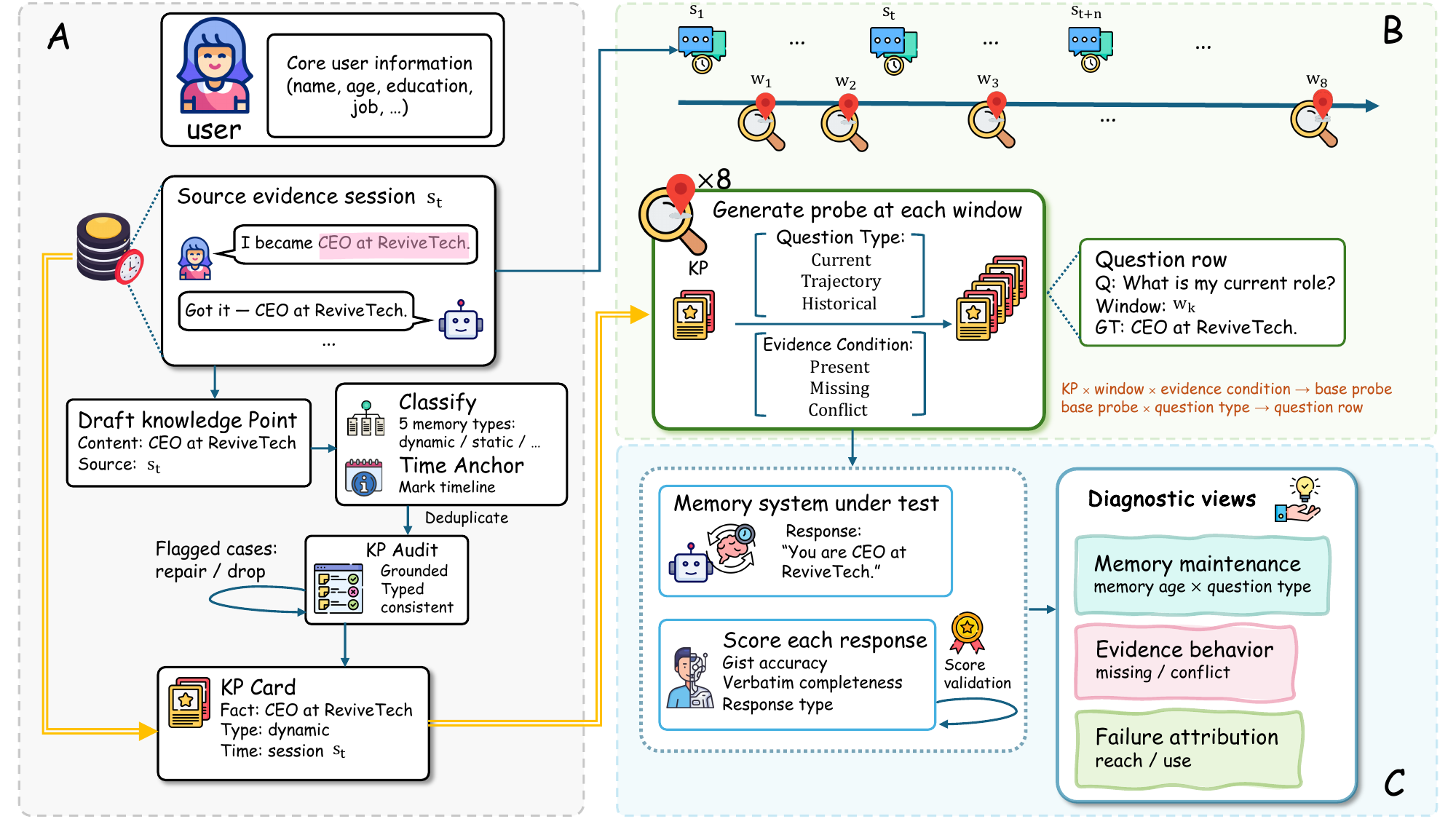}
\vspace{-2pt}

\caption{ 
Construction and evaluation schematic for \benchmark.
  \textbf{(A)} Source sessions are converted into typed knowledge points with
  session anchors and quality checks.
  \textbf{(B)} Probe construction pairs each knowledge point with a memory window
  and evidence condition, then expands the base probe by question type into
  question rows.
  \textbf{(C)} Memory-system responses are scored and summarized into diagnostic
  views of memory maintenance, evidence-condition behavior, and failure
  attribution.}
\label{fig:protocol}
\end{figure*}

\paragraph{Memory Benchmarks.} Memory evaluation spans long-context stress tests \citep{longbench,infinitebench,ruler}, newer long-context suites \citep{longbenchv2}, and long-term conversational memory \citep{msc,locomo,longmemeval}. Other benchmarks focus on personalization \citep{personamem,realmem}, agent memory \citep{memoryagentbench,beam}, dynamic profiles and evolving preferences \citep{horizonbench,memora}, stale or hallucinated memories \citep{stale,membench}, and missing or conflicting evidence \citep{halumem,memorybench}. These benchmarks broaden memory-system evaluation, but usually score question rows or interaction outcomes and aggregate across
them. \benchmark uses knowledge points as the analysis unit. Each fact is probed across windows, question types, and evidence conditions, so behavior is measured on the knowledge point rather than averaged across independent question rows.

\vspace{-5pt}
\section{The \benchmark Benchmark}
\label{sec:benchmark}
\vspace{-2pt}
Different from existing benchmarks that mainly evaluate isolated question rows or aggregate corpus-level accuracy, \benchmark uses the \emph{knowledge point} as the unit of measurement. A knowledge point is a typed fact about the user. For each knowledge point, we construct repeated probes that vary memory age, question type, and evidence condition while keeping the underlying fact fixed. This design allows us to test three behaviors that pooled QA scores usually merge: whether a fact remains usable as sessions accumulate, whether the system can answer current, historical, and trajectory questions about the same fact, and whether it behaves appropriately when evidence is present, missing, or contradicted. \Cref{fig:protocol} shows the full construction and evaluation flow: source sessions are converted into knowledge points, expanded into controlled probes, scored through memory-system responses, and summarized as diagnostic views. \Cref{tab:benchmark-comparison} positions this protocol against representative long-term and personalized-memory benchmarks \citep{locomo,longmemeval,personamem,halumem,memoryagentbench}.

\begin{table*}[t]
\centering
\footnotesize
\setlength{\tabcolsep}{3pt}
\renewcommand{\arraystretch}{0.95}
\begin{tabular}{@{}lccccc@{\hspace{5pt}}c@{}}
\toprule
Feature
& LoCoMo
& LongMemEval
& PersonaMem
& HaluMem
& MemoryAgentBench
& \textbf{\benchmark} \\
\midrule
\rowcolor{black!4}
\multicolumn{7}{@{}l}{\textit{General long-memory axes}} \\
\addlinespace[-1pt]
Multi-session users
& \yesmark & \yesmark & \yesmark & \yesmark & \yesmark & \textbf{\yesmark} \\
Personal facts/profiles
& \yesmark & \partmark & \yesmark & \yesmark & \partmark & \textbf{\yesmark} \\
Updates / change
& \partmark & \yesmark & \yesmark & \yesmark & \yesmark & \textbf{\yesmark} \\
Conflict / missing evidence
& \nomark & \partmark & \nomark & \yesmark & \yesmark & \textbf{\yesmark} \\
\midrule
\rowcolor{black!6}
\multicolumn{7}{@{}l}{\textbf{\textit{Knowledge-point protocol and attribution}}} \\
\addlinespace[-1pt]
Knowledge point as unit
& \nomark & \nomark & \nomark & \nomark & \nomark & \textbf{\yesmark} \\
Same fact across age windows
& \nomark & \nomark & \nomark & \partmark & \nomark & \textbf{\yesmark} \\
Same fact across question types
& \nomark & \nomark & \nomark & \nomark & \nomark & \textbf{\yesmark} \\
Evidence-condition probes
& \nomark & \partmark & \nomark & \yesmark & \yesmark & \textbf{\yesmark} \\
Failure attribution: reach vs. use
& \nomark & \nomark & \nomark & \nomark & \nomark & \textbf{\yesmark} \\
\bottomrule
\end{tabular}
\caption{Benchmark-design comparison. Rows are protocol axes, not quality
judgments; \yesmark{} = central; \partmark{} = related; \nomark{} = not part of
the main protocol.}
\label{tab:benchmark-comparison}
\end{table*}
\vspace{-4pt}
\subsection{Data Source and Knowledge Points}
\label{sec:benchmark-data}
\vspace{-2pt}

To evaluate a system along controlled dimensions, a per-fact protocol requires source data with two properties. First, facts must be grounded in multi-session user histories and anchored to specific sessions, so memory age can be defined. Second, the data must include distractor items for constructing missing-evidence and conflict probes, so evidence conditions can be tested. HaluMem-Medium \citep{halumem} provides both properties.
HaluMem evaluates whether memory systems hallucinate during extraction, updating, and question answering. However, its examples are not organized around a fixed knowledge point queried under multiple conditions. We therefore use HaluMem-Medium as source data and transform its histories, memory points, and diagnostic questions into knowledge-point probes. Substantive knowledge points are static facts, dynamic facts with earlier and updated states, or preference facts. Conflict and boundary distractors come from HaluMem's diagnostic questions and are reformulated for the per-fact protocol.

The benchmark covers 20 users and 835 typed knowledge points, expanded into 5{,}677 base probes, 15{,}422 question rows, and 200{,}453 scored answers across 13 memory-system configurations spanning four paradigms (Appendix~\ref{app:benchmark-details}). As shown in \Cref{tab:kp-distribution}, 635 knowledge points are substantive user facts and 200 are distractor knowledge points used for conflict and boundary probes. Each user contributes 35--48 knowledge points (mean 41.8).

\begin{table}[tb]
\centering
\footnotesize
\setlength{\tabcolsep}{5pt}
\renewcommand{\arraystretch}{0.95}
\begin{tabular}{lrr}
\toprule
Knowledge point group & Count & \% \\
\midrule
Static & 348 & 41.7 \\
Dynamic & 213 & 25.5 \\
Preference & 74 & 8.9 \\
Conflict distractor & 100 & 12.0 \\
Boundary distractor & 100 & 12.0 \\
\midrule
Total & 835 & 100.0 \\
\bottomrule
\end{tabular}
\caption{Distribution of knowledge points in \benchmark: static, dynamic, and
preference are substantive facts; conflict and boundary are distractors.}
\label{tab:kp-distribution}
\end{table}

\subsection{Probe Construction}
\label{sec:benchmark-probes}


Each probe in \benchmark reflects how long-term memory systems are queried in multi-session conversations. Rather than treating evaluation as a flat question-answering task, we fix the knowledge point and evaluate it along three dimensions that aggregate QA scores usually merge: how long the system can retain it (\emph{Memory age}), whether the system can flexibly utilize it across different question contexts (\emph{Question type}), and whether the system can handle it safely under unmentioned or misleading conditions (\emph{Evidence condition}).

Formally, a \emph{base probe} pairs a knowledge point, memory window $W_w$, and evidence condition, then expands into question rows. A memory window is an evaluation checkpoint; its \emph{prefix} is the conversation history shown to the system. If $W_w$ ends after session $t_{\mathrm{eval}}$, the system receives sessions with indices $<t_{\mathrm{eval}}$. This differs from boundary probes, which query unmentioned facts.
\paragraph{Memory age.}
To track how memory persists as the conversation grows, we evaluate each user's history at eight chronological checkpoints (rather than on disjoint session blocks) to map a continuous retention trace. The evaluation checkpoints are chosen to span from early to saturated histories, with extra windows anchored near dynamic updates to probe pre- and post-update states. A knowledge point is only probed after its initial appearance. If it first appears in session $t_{\mathrm{source}}$, its memory age at $W_w$ is $t_{\mathrm{eval}} - t_{\mathrm{source}}$ sessions.

\paragraph{Question type.}
Question type tests whether the same fact supports different uses. Users ask beyond the current state of a fact. Each substantive knowledge point is instantiated as a \qcurrent question (what is the
current state?), a \qhistorical question (what was the state at an earlier
point?), and a \qtrajectory question (how has the state changed over
time?); distractor knowledge points, which drive the evidence-condition
probes described next, carry only \qcurrent\ and \qhistorical\ questions.
\paragraph{Evidence condition.}
Evidence condition tests how systems respond when evidence is absent or
conflicting.
Real conversations include queries about unmentioned facts (missing evidence) and queries whose premise contradicts memory (conflicting evidence). The standard probes provide the evidence-present case. We therefore expand distractor knowledge points into two additional probe families:
\emph{boundary probes}, which query unmentioned facts, and \emph{conflict
probes}, which assert a false premise that contradicts memory.
Both families are adapted from HaluMem's diagnostic taxonomy and aligned with
the per-fact unit.
\begin{table*}[t]
\centering
\begingroup
\scriptsize
\setlength{\tabcolsep}{0pt}
\renewcommand{\arraystretch}{0.98}
\begin{tabular}{@{}>{\raggedright\arraybackslash}p{3.20cm}@{\hspace{2pt}}%
>{\centering\arraybackslash}p{1.03cm}@{}%
>{\centering\arraybackslash}p{1.03cm}@{}%
>{\centering\arraybackslash}p{1.03cm}@{\hspace{3pt}}%
>{\centering\arraybackslash}p{1.03cm}@{}%
>{\centering\arraybackslash}p{1.03cm}@{}%
>{\centering\arraybackslash}p{1.03cm}@{\hspace{3pt}}%
>{\centering\arraybackslash}p{1.03cm}@{}%
>{\centering\arraybackslash}p{1.03cm}@{}%
>{\centering\arraybackslash}p{1.03cm}@{\hspace{3pt}}%
>{\centering\arraybackslash}p{1.03cm}@{}%
>{\centering\arraybackslash}p{1.03cm}@{}%
>{\centering\arraybackslash}p{1.03cm}@{}}
\toprule
& \multicolumn{3}{c}{Current} & \multicolumn{3}{c}{Historical} & \multicolumn{3}{c}{Trajectory} & \multicolumn{3}{c}{Overall} \\
\cmidrule(lr){2-4}\cmidrule(lr){5-7}\cmidrule(lr){8-10}\cmidrule(l){11-13}
System &
\multicolumn{1}{c}{Fresh} & \multicolumn{1}{c}{Saturated} & \multicolumn{1}{c}{$\Delta$Forget} &
\multicolumn{1}{c}{Fresh} & \multicolumn{1}{c}{Saturated} & \multicolumn{1}{c}{$\Delta$Forget} &
\multicolumn{1}{c}{Fresh} & \multicolumn{1}{c}{Saturated} & \multicolumn{1}{c}{$\Delta$Forget} &
\multicolumn{1}{c}{Fresh} & \multicolumn{1}{c}{Saturated} & \multicolumn{1}{c}{$\Delta$Forget} \\
\midrule
\rowcolor{black!4}\multicolumn{13}{c}{\textit{Long Context}} \\
Gemini-3-Flash & \cellcolor{blue!18}66.0 & \cellcolor{blue!12}39.7 & 26.2 & \cellcolor{blue!12}59.6 & \cellcolor{blue!18}47.1 & 12.5 & \cellcolor{blue!12}31.5 & \cellcolor{blue!7}11.0 & 20.5 & \cellcolor{blue!12}52.3 & \cellcolor{blue!12}32.6 & 19.7 \\
Qwen3.5-35B & \cellcolor{blue!18}68.8 & \cellcolor{blue!3}32.7 & 36.1 & \cellcolor{blue!18}66.2 & \cellcolor{blue!12}44.2 & 21.9 & \cellcolor{blue!18}\textbf{49.0} & \cellcolor{white}6.7 & 42.3 & \cellcolor{blue!18}\textbf{61.3} & \cellcolor{blue!7}27.9 & 33.5 \\
GPT-5-nano & \cellcolor{blue!18}67.8 & \cellcolor{blue!7}34.8 & 33.0 & \cellcolor{blue!12}62.4 & \cellcolor{blue!12}41.9 & 20.5 & \cellcolor{blue!12}38.4 & \cellcolor{white}6.5 & 31.9 & \cellcolor{blue!18}56.2 & \cellcolor{blue!7}27.7 & 28.5 \\
\addlinespace[1pt]
\rowcolor{black!4}\multicolumn{13}{c}{\textit{RAG}} \\
HippoRAG-v2 & \cellcolor{blue!18}\textbf{69.7} & \cellcolor{blue!18}\textbf{45.4} & 24.4 & \cellcolor{blue!18}\textbf{67.6} & \cellcolor{blue!18}\textbf{50.9} & 16.7 & \cellcolor{blue!12}33.0 & \cellcolor{blue!7}13.4 & 19.7 & \cellcolor{blue!18}56.8 & \cellcolor{blue!18}\textbf{36.5} & 20.2 \\
text-emb-3-small & \cellcolor{blue!12}65.7 & \cellcolor{blue!18}43.4 & 22.3 & \cellcolor{blue!12}59.3 & \cellcolor{blue!18}47.5 & 11.8 & \cellcolor{blue!3}19.7 & \cellcolor{blue!3}9.4 & 10.3 & \cellcolor{blue!7}48.2 & \cellcolor{blue!18}33.4 & 14.8 \\
Qwen3-Emb & \cellcolor{blue!7}61.2 & \cellcolor{blue!12}41.1 & 20.1 & \cellcolor{blue!12}59.2 & \cellcolor{blue!18}48.2 & 11.0 & \cellcolor{white}15.7 & \cellcolor{blue!3}8.6 & 7.1 & \cellcolor{blue!7}45.4 & \cellcolor{blue!12}32.6 & 12.7 \\
BM25 & \cellcolor{blue!7}61.2 & \cellcolor{blue!12}39.9 & 21.3 & \cellcolor{blue!12}56.2 & \cellcolor{blue!18}46.9 & 9.3 & \cellcolor{blue!3}18.3 & \cellcolor{blue!3}9.2 & 9.1 & \cellcolor{blue!7}45.2 & \cellcolor{blue!12}32.0 & 13.2 \\
\addlinespace[1pt]
\rowcolor{black!4}\multicolumn{13}{c}{\textit{External Memory}} \\
SimpleMem & \cellcolor{blue!7}60.4 & \cellcolor{blue!18}43.8 & 16.6 & \cellcolor{blue!12}58.0 & \cellcolor{blue!12}44.8 & 13.2 & \cellcolor{blue!12}40.1 & \cellcolor{blue!18}17.3 & 22.8 & \cellcolor{blue!12}52.9 & \cellcolor{blue!18}35.3 & 17.5 \\
\rowcolor{black!2}\emph{SimpleMem (gpt-4.1-mini)} & 63.7 & 40.9 & 22.8 & 65.7 & 41.2 & 24.6 & 42.7 & 13.4 & 29.4 & 57.4 & 31.8 & 25.6 \\
REMem & \cellcolor{blue!12}62.1 & \cellcolor{blue!18}43.1 & 18.9 & \cellcolor{blue!7}53.7 & \cellcolor{blue!12}43.1 & 10.6 & \cellcolor{blue!7}24.1 & \cellcolor{blue!7}12.6 & 11.5 & \cellcolor{blue!7}46.6 & \cellcolor{blue!18}33.0 & 13.7 \\
AMem & \cellcolor{blue!3}55.3 & \cellcolor{blue!18}42.0 & 13.3 & \cellcolor{blue!3}47.7 & \cellcolor{blue!7}39.3 & 8.5 & \cellcolor{white}15.6 & \cellcolor{blue!7}10.9 & 4.7 & \cellcolor{blue!3}39.5 & \cellcolor{blue!12}30.7 & 8.8 \\
Mem0 & \cellcolor{white}50.1 & \cellcolor{blue!3}31.5 & 18.7 & \cellcolor{white}37.5 & \cellcolor{blue!3}29.7 & 7.7 & \cellcolor{white}9.6 & \cellcolor{blue!3}8.2 & 1.5 & \cellcolor{white}32.4 & \cellcolor{blue!3}23.1 & 9.3 \\
\addlinespace[1pt]
\rowcolor{black!4}\multicolumn{13}{c}{\textit{Agentic Memory}} \\
Mem-T & \cellcolor{blue!7}58.9 & \cellcolor{blue!12}40.4 & 18.5 & \cellcolor{blue!12}59.8 & \cellcolor{blue!18}47.3 & 12.5 & \cellcolor{blue!18}47.2 & \cellcolor{blue!18}\textbf{19.8} & 27.4 & \cellcolor{blue!18}55.3 & \cellcolor{blue!18}35.8 & 19.5 \\
MIRIX & \cellcolor{blue!3}56.4 & \cellcolor{white}26.4 & 30.0 & \cellcolor{white}40.5 & \cellcolor{white}23.8 & 16.7 & \cellcolor{white}12.4 & \cellcolor{white}4.8 & 7.6 & \cellcolor{white}36.4 & \cellcolor{white}18.3 & 18.1 \\
\rowcolor{black!2}\emph{MIRIX (gpt-4.1-mini)} & 53.2 & 36.2 & 17.0 & 69.3 & 46.3 & 23.0 & 29.3 & 16.1 & 13.2 & 50.6 & 32.9 & 17.7 \\
\bottomrule
\end{tabular}
\endgroup

\caption{\textbf{Memory maintenance by question type.} Fresh averages Gist
over checkpoints W1 and W2, Saturated averages Gist over checkpoints W7 and
W8, and $\Delta$Forget is their gap computed before rounding, all in percentage points. Score cells are
shaded within each column; bold marks each
Fresh/Saturated leader among the main rows. Gray rows are paper-native backbone
sensitivity rows and are not counted as additional main systems.}
\label{tab:main-leaderboard}
\end{table*}
\vspace{-10pt}
\subsection{Metrics}
\vspace{-2pt}
A single accuracy metric would conflate distinct response behaviors. Each response is therefore assigned a score tuple $(g,v,r)$. Here, $g$ is binary \emph{Gist accuracy}, which captures semantic correctness; $v$ is continuous \emph{Verbatim completeness} in $[0,1]$, which checks whether canonical answer tokens appear; and $r$ is \emph{Response type}, which supports diagnostics over correct answers, abstentions, and hallucinations. GPT-4o is the primary judge; Gemini-3-Flash checks reliability on a stratified 200-probe sample under the same rubric (Appendix~\ref{app:judge-reliability}).
\vspace{-4pt}
\subsection{Diagnostic Views} 
\vspace{-2pt}
\label{sec:benchmark-diagnostics}


We report three diagnostic views instead of one pooled score: memory maintenance, evidence-condition behavior, and reach/use attribution.
\vspace{-4pt}
\paragraph{Memory maintenance.}
For each knowledge point and question type, we aggregate Gist accuracy
across all eight windows to ask whether memory persists as the conversation
grows. We report this in two ways. The first is the full retention trace
$W_1,\dots,W_8$, visualized as a per-system retention curve. The second is
three scalar summaries: \fresh\ accuracy ($W_1, W_2$), measuring recent
access; \saturated\ accuracy ($W_7, W_8$), measuring aged-memory retention;
and \deltaforget, the gap between the two. \deltaforget\ is a diagnostic
gap, not a standalone leaderboard: a small value can mean either stable
retention or uniformly low endpoint accuracy.
\vspace{-4pt}
\paragraph{Evidence-condition behavior.}
For each system, we aggregate response-type rates over distractor probes: correct boundary refusal, correct conflict resolution, and hallucination on both. Safe abstention and false-premise rejection differ, but pooled accuracy treats both as wrong.

\vspace{-4pt}
\paragraph{Failure attribution.}
When a system answers incorrectly, we ask whether the gold evidence was unreachable or reachable but unused. This diagnostic runs on dedicated samples rather than the full benchmark and has three components.

\emph{Oracle on hard failures.} We supply gold memory evidence directly to probes that all systems respond incorrectly and then re-score. The lift over the production baseline tests whether these probes can be solved when the needed evidence is explicit. This open-book check is not a retrieval setting; it only checks whether the hard probes are unanswerable.

\emph{Reach-vs.-use replay.} On a broader sample, a simple
Text-emb-3-small retriever checks whether it can reach the source session containing the gold evidence. We combine this reach signal with the original production answer to separate reach misses from reached-but-unsolved cases.

\emph{Oracle on retriever-reached but unsolved probes.} We then rerun the open-book oracle on the retriever-reached but unsolved cases. If accuracy rises, the error is recoverable with explicit evidence; if not, the case may remain unsolved even with evidence.

\section{Experiment and Findings}
\vspace{-4pt}
\label{sec:experiments}
This section evaluates \benchmark along the diagnostic views defined in \cref{sec:benchmark-diagnostics}. We then report memory maintenance across sessions and question types, behavior under missing and conflicting evidence, and failure attribution by retrieval reach and evidence use. We close with the main implications for long-term memory system design.
\vspace{-4pt}

\begin{figure*}[t]
\centering
\includegraphics[width=0.85\textwidth]{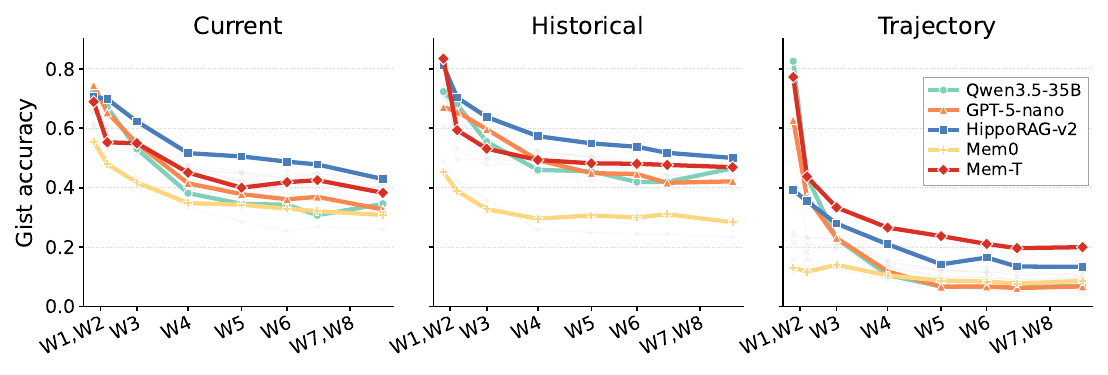}
\vspace{-5pt}
\caption{\textbf{Memory-age traces.} Gist traces from checkpoint W1 through
W8 for the main configurations. Each panel fixes one question type: current
state, earlier state, or trajectory of change. Gray lines show the remaining
systems, and colored lines highlight representative configurations used in the text.
Endpoints are summarized in \Cref{tab:main-leaderboard}.}
\label{fig:window-retention}
\end{figure*}
\vspace{-5pt}

\subsection{Experiment Setting}
\label{sec:experiment-setting}
\vspace{-2pt}
We evaluate 13 memory-system configurations across four paradigms.
\textbf{(a) Long-context configurations} include Qwen3.5-35B~\citep{qwen3},
Gemini-3-Flash~\citep{gemini25}, and GPT-5-nano~\citep{oai_gpt5nano}; they
read visible history directly in the model context.
\textbf{(b) RAG configurations} include BM25~\citep{bm25},
Text-emb-3-small~\citep{oai_emb3}, Qwen3-Emb~\citep{qwen3emb}, and HippoRAG-v2~\citep{hipporag}; they retrieve memory evidence before generation. \textbf{(c) External-memory configurations} include Mem0~\citep{mem0},
SimpleMem~\citep{simplemem}, REMem~\citep{remem}, and AMem~\citep{amem}; they
maintain memory stores. \textbf{(d) Agentic-memory configurations} include
MIRIX~\citep{mirix} and Mem-T~\citep{memt}; they use policy-driven or multi-agent memory management. Appendix~\ref{app:system-configurations} gives the full configuration for all 13 systems.

For long-context models, each probe is answered by prompting the model with the
visible session prefix for that window. For the other configurations, the
memory state for each checkpoint is built or updated only from the same visible
prefix; future sessions are not available to storage, retrieval, or generation.
The system then returns retrieved or stored memory evidence for the probe. We
pass that evidence to a shared gpt-4o-mini answer generator with the same
answer prompt, so the main comparison changes the memory mechanism while
holding the final generator fixed. We also report paper-native gpt-4.1-mini
sensitivity rows for SimpleMem and MIRIX where the original systems specify
that backbone. Unless otherwise stated, results use Gist accuracy as the
correctness signal.
The following subsections use this setup to test the three diagnostic views introduced in \Cref{sec:benchmark-diagnostics}.

\vspace{-2pt}
\subsection{Memory Maintenance Across Sessions and Question Types}
\label{sec:final-accuracy}
\textbf{Pooled accuracy hides when and how memory fails.}
A pooled final-accuracy score removes the two variables that define long-term memory use: when a knowledge point is queried and what the question asks the system to do with it. \benchmark therefore reports \fresh access, averaged over W1 and W2; \saturated retention, averaged over W7 and W8; and \deltaforget, their gap. We compute these summaries separately for current, historical, and trajectory questions. \Cref{tab:main-leaderboard} reports these summaries for all main configurations, and \Cref{fig:window-retention} shows the traces from W1 through W8.
\vspace{-2pt}
These results show why a single leaderboard is lossy. On overall \saturated Gist, HippoRAG-v2 has the strongest endpoint (36.5\%), but the leader changes by question type: HippoRAG-v2 leads saturated current and historical questions (45.4\% and 50.9\%), while Mem-T leads saturated trajectory questions (19.8\%). Trajectory is therefore not just harder lookup; it asks for a different memory behavior. Long-context models fail differently: Qwen3.5-35B and GPT-5-nano have high \fresh trajectory scores (49.0\% and 38.4\%), but fall to 6.7\% and 6.5\% when saturated.
Appendix~\ref{app:full-results} reports the corresponding bootstrap rank checks in \Cref{tab:appendix-rank-robustness}.
\vspace{-2pt}
This trajectory drop suggests that relevant evidence alone is not enough. A trajectory answer must connect multiple states of the same knowledge point, identify their temporal order, and express the update. Long-context models receive the visible conversation history, but as it grows, old and new mentions compete with later sessions; the model may answer a local current or historical query while failing to organize the temporal relation. RAG systems show a complementary limitation: HippoRAG-v2 leads saturated current and historical questions, but reaches only 13.4\% on saturated trajectory. In both cases, the hard part is using multiple states as a temporal update trace, not recalling one state. A small \deltaforget can still hide low performance at both endpoints, so we interpret the gap together with \fresh and \saturated.


\subsection{Behavior Under Missing and Conflicting Evidence}
\label{sec:evidence-condition-results}

\begin{figure}[tb]
\centering
\includegraphics[width=0.98\columnwidth]{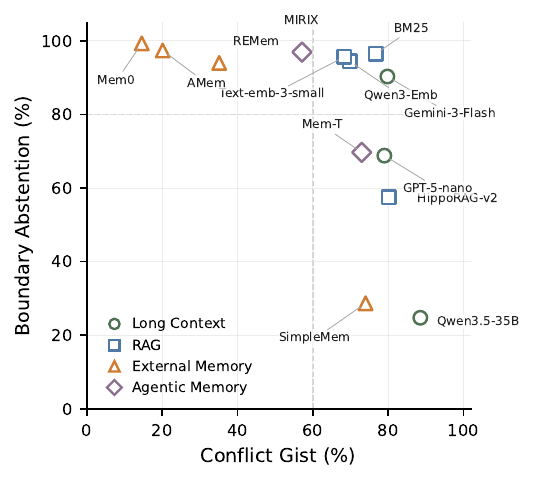}
\vspace{-5pt}
\caption{\textbf{Conflict Gist versus boundary abstention by system.} The two
axes separate missing-evidence refusal from false-premise resolution; dashed
guides mark the high-boundary and high-conflict regions.}
\label{fig:conflict-boundary-scatter}
\end{figure}

\begin{table}[tb]
\centering
\begingroup
\setlength{\tabcolsep}{1.5pt}
\renewcommand{\arraystretch}{0.95}
\footnotesize
\begin{tabular*}{\columnwidth}{@{\extracolsep{\fill}}lcccccc@{}}
\toprule
 & \multicolumn{3}{c}{Conflict} & \multicolumn{3}{c}{Boundary} \\
\cmidrule(lr){2-4}\cmidrule(l){5-7}
System & Gist & Abst. & Hallu. & Gist & Abst. & Hallu. \\
\midrule
\rowcolor{black!4}\multicolumn{7}{c}{\textit{Long Context}} \\
Qwen3.5-35B & 88.6 & 8.4 & 3.9 & 89.6 & 24.8 & 10.4 \\
Gemini-3-Flash & 79.8 & 15.2 & 6.4 & 96.0 & 90.3 & 3.8 \\
GPT-5-nano & 79.0 & 16.9 & 5.7 & 95.1 & 68.8 & 4.9 \\
\addlinespace[1pt]
\rowcolor{black!4}\multicolumn{7}{c}{\textit{RAG}} \\
HippoRAG-v2 & 80.2 & 11.7 & 11.1 & 94.4 & 57.5 & 5.2 \\
BM25 & 76.7 & 20.3 & 3.6 & 96.6 & 96.4 & 3.4 \\
Qwen3-Emb & 69.8 & 23.1 & 8.0 & 95.8 & 94.5 & 4.2 \\
Text-emb-3-small & 68.3 & 25.9 & 6.9 & 96.4 & 95.6 & 3.6 \\
\addlinespace[1pt]
\rowcolor{black!4}\multicolumn{7}{c}{\textit{External Memory}} \\
SimpleMem & 74.0 & 13.9 & 12.6 & 91.6 & 28.7 & 7.6 \\
REMem & 35.1 & 61.8 & 4.9 & 94.5 & 94.0 & 3.6 \\
AMem & 20.1 & 78.3 & 1.1 & 97.4 & 97.4 & 2.6 \\
Mem0 & 14.6 & 82.5 & 2.7 & 99.3 & 99.3 & 0.7 \\
\addlinespace[1pt]
\rowcolor{black!4}\multicolumn{7}{c}{\textit{Agentic Memory}} \\
Mem-T & 73.0 & 18.3 & 11.4 & 91.1 & 69.7 & 8.6 \\
MIRIX & 57.1 & 27.5 & 18.1 & 97.1 & 96.9 & 2.9 \\
\bottomrule
\end{tabular*}
\endgroup

\caption{\textbf{Conflict and boundary profile.} For both conflict and boundary probes we report Gist accuracy, abstention rate (Abst.), and hallucination rate (Hallu.). On conflict probes, Gist measures whether the system resolves a false
premise using memory; on boundary probes, abstention measures safe refusal when the requested fact is missing. All values are percentages.}

\label{tab:conflict-boundary-profile}
\end{table}

\textbf{Abstaining is not resolving.}
Reliable memory systems must handle missing evidence and false premises, not only recall supported facts. Boundary probes ask about facts absent from the history and should be refused. Conflict probes contain a false premise and require correction from memory. \Cref{tab:conflict-boundary-profile} and \Cref{fig:conflict-boundary-scatter} compare the two axes.
These two behaviors are empirically separate. Mem0, AMem, and REMem are boundary-safe: their boundary abstention rates are 99.3\%, 97.4\%, and 94.0\%,
but they rarely resolve conflict probes. Their conflict Gist scores are
14.6\%, 20.1\%, and 35.1\%. Their conflict failures are mostly abstentions
rather than fabrications: conflict hallucination is 2.7\%, 1.1\%, and 4.9\%.
This pattern suggests an evidence-use failure at the memory-to-answer interface, not a lack of stored memory. Under the main answer backbone, boundary probes can be handled by abstention when support is missing, while conflict probes require the generator to use memory against the user's false premise. Relevant memory evidence is present, but the system must first detect that the premise contradicts memory and then convert the memory into a correction. Thus, these systems appear safe on missing evidence while often treating contradicted evidence as if it were absent. Appendix~\ref{app:backbone-sensitivity} reports answer-backbone ablations showing that conflict resolution is sensitive to how memory evidence is passed to the generator.
\vspace{-8pt}
\subsection{Failure Attribution: Retrieval Reach vs. Evidence Use}
\label{sec:failure-origin-results}
\textbf{Reaching evidence is not the same as using it.}
We analyze failure origin in three steps. First, we check whether hard probes can be answered when gold evidence is given
directly. Second, we check whether a simple retriever can reach the gold evidence. Third, for cases where evidence is reachable but the system still fails, we give the gold evidence and test whether the answer can be recovered.

\begin{figure}[tb]
\centering
\includegraphics[width=0.96\columnwidth]{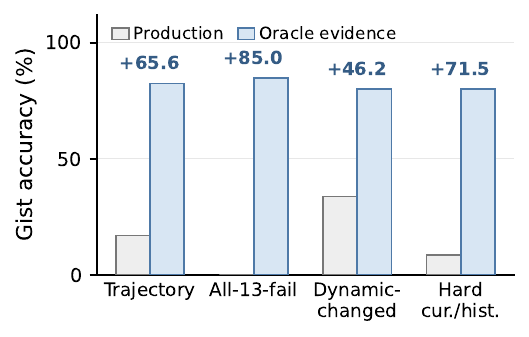}
\vspace{-5pt}
\caption{\textbf{Failure-attribution decomposition.} The oracle component gives the gold evidence directly to test whether hard probes are answerable; it is not a retrieval method. The retrieval replay splits failures into reach vs. reached-but-unsolved. }
\label{fig:oracle-recovery-decomposition}
\end{figure}

We first analyze 120 hard probes: 40 all-systems-fail trajectory probes, 40 dynamic-changed trajectory probes, and 40 hard current/historical control probes. We supply gold memory evidence directly to the generator, an open-book setting that tests answerability rather than retrieval. Oracle Gist rises to 80--85\% in every bucket, while production baselines stay between 0\% and 33.8\% (\Cref{fig:oracle-recovery-decomposition}). Thus many hard probes are answerable with the right evidence; we next ask whether that evidence was unreachable or reachable but unused.

To check whether evidence is reachable, we run a 300-probe replay with a simple Text-emb-3-small retriever. We record two signals per probe: $R{=}1$ if the retriever reaches the gold session, and $U{=}1$ if the original production answer solves the probe. Of the 300 probes, 21 are reach misses (7.0\%), 220 are retriever-reached but unsolved (73.3\%), and 59 are solved (19.7\%). Thus the dominant failure case is $R{=}1,U{=}0$ (\Cref{tab:recovery-reachable-unsolved}). Extending this reach/use replay to all 13 main configurations gives the same picture: $P(U{=}0\mid R{=}1)$ ranges from 69.2\% for HippoRAG-v2 to 88.2\% for MIRIX. We then ask whether these reached but unsolved probes become answerable with explicit evidence.

\begin{table}[tb]
\centering
\begingroup
\setlength{\tabcolsep}{1.5pt}
\renewcommand{\arraystretch}{0.95}
\footnotesize
\begin{tabular*}{\columnwidth}{@{\extracolsep{\fill}}lccc@{}}
\toprule
System & $R{=}1,U{=}0$ & $P(U{=}0\mid R{=}1)$ & Oracle Gist \\
\midrule
\rowcolor{black!4}\multicolumn{4}{c}{\textit{Long Context}} \\
Qwen3.5-35B & 232/279 & 83.2 & 81.5 \\
Gemini-3-Flash & 224/279 & 80.3 & 80.4 \\
GPT-5-nano & 225/279 & 80.6 & 80.9 \\
\addlinespace[1pt]
\rowcolor{black!4}\multicolumn{4}{c}{\textit{RAG}} \\
HippoRAG-v2 & 193/279 & 69.2 & 83.4 \\
BM25 & 231/279 & 82.8 & 81.8 \\
Qwen3-Emb & 222/279 & 79.6 & 81.5 \\
Text-emb-3-small & 220/279 & 78.9 & 80.9 \\
\addlinespace[1pt]
\rowcolor{black!4}\multicolumn{4}{c}{\textit{External Memory}} \\
SimpleMem & 208/279 & 74.6 & 82.7 \\
REMem & 211/279 & 75.6 & 83.9 \\
AMem & 237/279 & 84.9 & 82.3 \\
Mem0 & 234/279 & 83.9 & 80.8 \\
\addlinespace[1pt]
\rowcolor{black!4}\multicolumn{4}{c}{\textit{Agentic Memory}} \\
Mem-T & 220/279 & 78.9 & 82.3 \\
MIRIX & 246/279 & 88.2 & 81.7 \\
\bottomrule
\end{tabular*}
\endgroup

\caption{\textbf{Retriever-reached but unsolved failures.} The $R{=}1,U{=}0$
column gives the count of retriever-reached but unsolved probes over the 279
retriever-reached probes; $P(U{=}0\mid R{=}1)$ and Oracle Gist are percentages.
Oracle Gist re-supplies the gold memory evidence on each system's $R{=}1,U{=}0$
subset (per-row denominator) and re-scores with gpt-4o-mini backbone.}
\label{tab:recovery-reachable-unsolved}
\end{table}

Finally, we rerun the open-book oracle on each main system's $R{=}1,U{=}0$ probes (Oracle Gist column, \Cref{tab:recovery-reachable-unsolved}). All 13 systems recover to 80.4\%--83.9\%, a pooled lift of 81.8 percentage points over the 0\% baseline. This confirms that most errors are not caused by unreachable evidence. Reaching the gold session is only a weak precondition: it does not guarantee that the relevant span, temporal relation, or conflict signal is exposed to the generator. The harder step is selecting and presenting reachable evidence in a form the generator can use, especially when the answer requires comparing states over time or correcting a false premise.

\vspace{-6pt}
\subsection{Summary and Implications}
\vspace{-2pt}

\label{sec:findings-summary}
Taken together, the results suggest that long-term memory failure is not merely a storage or retrieval failure, but a temporally grounded evidence-use failure. Long-context systems read the visible history directly, but their aged-trajectory collapse is consistent with attention dilution and weak temporal organization: old and new mentions must compete with many irrelevant sessions before the model can form an update. RAG systems recover current and historical evidence more effectively, but retrieval exposes local evidence pieces rather than a coherent update path across states. External-memory systems can appear safe on missing evidence because conservative abstention is sufficient there, but conflict probes require active correction: rejecting the false premise while committing to the remembered fact. The reach/use audit connects these patterns: many wrong answers occur after relevant evidence is already reachable. Personalized memory systems therefore need more than larger context windows, stronger retrievers, or safer abstention; they need mechanisms that expose reachable evidence with temporal and conflict structure for multi-session queries.

\vspace{-6pt}
\section{Conclusion}
\vspace{-2pt}
\benchmark recasts long-term memory evaluation around the knowledge point:
rather than pooling question rows, it holds a single user fact fixed and varies
its memory age, the question asked about it, and the evidence available. This
makes visible what an aggregate score cannot: when a fact decays over sessions,
when a system recalls a state but not how it changed, and when it abstains
safely yet fails to correct a false premise. Across 13 memory-system
configurations and four paradigms, systems with similar pooled accuracy diverge
along these axes, and most errors persist even when the relevant evidence is
already reachable. Long-term memory is therefore limited less by storing or
retrieving facts than by using reachable evidence during inference.

\section*{Limitations}
\benchmark is derived from HaluMem-Medium and covers 20 users from a single source distribution, so its results should not be read as spanning all domains or interaction styles. Our main numbers compare end-to-end configurations rather than isolated memory mechanisms: long-context rows are provider-native models, while most other rows share a gpt-4o-mini generator but differ in their retrievers and embeddings (AMem, for instance, keeps its original all-MiniLM-L6-v2 embedding). Because backbone choice alone can shift a system across much of the leaderboard, cross-paradigm rankings should be read at the configuration level. The reach/use attribution further relies on a single Text-emb-3-small retriever and an open-book oracle over hard probes, so the roughly 10$\times$ imbalance is a directional finding whose magnitude depends on the retriever. Lastly, scoring relies on an LLM judge, with reliability checked by a second judge on a stratified sample.

\bibliography{references}

@misc{hu2026memoryageaiagents,
      title={Memory in the Age of AI Agents}, 
      author={Yuyang Hu and Shichun Liu and Yanwei Yue and Guibin Zhang and Boyang Liu and Fangyi Zhu and Jiahang Lin and Honglin Guo and Shihan Dou and Zhiheng Xi and Senjie Jin and Jiejun Tan and Yanbin Yin and Jiongnan Liu and Zeyu Zhang and Zhongxiang Sun and Yutao Zhu and Hao Sun and Boci Peng and Zhenrong Cheng and Xuanbo Fan and Jiaxin Guo and Xinlei Yu and Zhenhong Zhou and Zewen Hu and Jiahao Huo and Junhao Wang and Yuwei Niu and Yu Wang and Zhenfei Yin and Xiaobin Hu and Yue Liao and Qiankun Li and Kun Wang and Wangchunshu Zhou and Yixin Liu and Dawei Cheng and Qi Zhang and Tao Gui and Shirui Pan and Yan Zhang and Philip Torr and Zhicheng Dou and Ji-Rong Wen and Xuanjing Huang and Yu-Gang Jiang and Shuicheng Yan},
      year={2026},
      eprint={2512.13564},
      archivePrefix={arXiv},
      primaryClass={cs.CL},
      url={https://arxiv.org/abs/2512.13564}, 
}

@misc{huang2026rethinkingmemorymechanismsfoundation,
      title={Rethinking Memory Mechanisms of Foundation Agents in the Second Half: A Survey}, 
      author={Wei-Chieh Huang and Weizhi Zhang and Yueqing Liang and Yuanchen Bei and Yankai Chen and Tao Feng and Xinyu Pan and Zhen Tan and Yu Wang and Tianxin Wei and Shanglin Wu and Ruiyao Xu and Liangwei Yang and Rui Yang and Wooseong Yang and Chin-Yuan Yeh and Hanrong Zhang and Haozhen Zhang and Siqi Zhu and Henry Peng Zou and Wanjia Zhao and Song Wang and Wujiang Xu and Zixuan Ke and Zheng Hui and Dawei Li and Yaozu Wu and Langzhou He and Chen Wang and Xiongxiao Xu and Baixiang Huang and Juntao Tan and Shelby Heinecke and Huan Wang and Caiming Xiong and Ahmed A. Metwally and Jun Yan and Chen-Yu Lee and Hanqing Zeng and Yinglong Xia and Xiaokai Wei and Ali Payani and Yu Wang and Haitong Ma and Wenya Wang and Chenguang Wang and Yu Zhang and Xin Wang and Yongfeng Zhang and Jiaxuan You and Hanghang Tong and Xiao Luo and Xue Liu and Yizhou Sun and Wei Wang and Julian McAuley and James Zou and Jiawei Han and Philip S. Yu and Kai Shu},
      year={2026},
      eprint={2602.06052},
      archivePrefix={arXiv},
      primaryClass={cs.CL},
      url={https://arxiv.org/abs/2602.06052}, 
}

@misc{halumem,
  title = {HaluMem: Evaluating Hallucinations in Memory Systems of Agents},
  author = {Chen, Ding and Niu, Simin and Li, Kehang and Liu, Peng and Zheng, Xiangping and Tang, Bo and Li, Xinchi and Xiong, Feiyu and Li, Zhiyu},
  year = {2026},
  eprint = {2511.03506},
  archivePrefix = {arXiv},
  primaryClass = {cs.CL}
}

@misc{memora,
  title = {From Recall to Forgetting: Benchmarking Long-Term Memory for Personalized Agents},
  author = {Uddin, Md Nayem and Shubham, Kumar and Blanco, Eduardo and Baral, Chitta and Wang, Gengyu},
  year = {2026},
  eprint = {2604.20006},
  archivePrefix = {arXiv},
  primaryClass = {cs.CL}
}

@misc{mem2actbench,
  title = {Mem2ActBench: A Benchmark for Evaluating Long-Term Memory Utilization in Task-Oriented Autonomous Agents},
  author = {Shen, Yiting and Li, Kun and Zhou, Wei and Hu, Songlin},
  year = {2026},
  eprint = {2601.19935},
  archivePrefix = {arXiv},
  primaryClass = {cs.CL}
}

@misc{beam,
  title = {Beyond a Million Tokens: Benchmarking and Enhancing Long-Term Memory in LLMs},
  author = {Tavakoli, Mohammad and Salemi, Alireza and Ye, Carrie and Abdalla, Mohamed and Zamani, Hamed and Mitchell, J. Ross},
  year = {2026},
  eprint = {2510.27246},
  archivePrefix = {arXiv},
  primaryClass = {cs.CL}
}

@inproceedings{locomo,
  title = {Evaluating Very Long-Term Conversational Memory of {LLM} Agents},
  author = {Maharana, Adyasha and Lee, Dong-Ho and Tulyakov, Sergey and Bansal, Mohit and Barbieri, Francesco and Fang, Yuwei},
  booktitle = {Proceedings of the 62nd Annual Meeting of the Association for Computational Linguistics (Volume 1: Long Papers)},
  pages = {13851--13870},
  year = {2024},
  publisher = {Association for Computational Linguistics},
  doi = {10.18653/v1/2024.acl-long.747}
}

@inproceedings{longmemeval,
  title = {LongMemEval: Benchmarking Chat Assistants on Long-Term Interactive Memory},
  author = {Wu, Di and Wang, Hongwei and Yu, Wenhao and Zhang, Yuwei and Chang, Kai-Wei and Yu, Dong},
  booktitle = {International Conference on Learning Representations},
  year = {2025}
}

@misc{personamem,
  title = {Know Me, Respond to Me: Benchmarking {LLM}s for Dynamic User Profiling and Personalized Responses at Scale},
  author = {Jiang, Bowen and Hao, Zhuoqun and Cho, Young-Min and Li, Bryan and Yuan, Yuan and Chen, Sihao and Ungar, Lyle and Taylor, Camillo J. and Roth, Dan},
  year = {2025},
  eprint = {2504.14225},
  archivePrefix = {arXiv},
  primaryClass = {cs.CL}
}

@misc{memoryagentbench,
  title = {Evaluating Memory in {LLM} Agents via Incremental Multi-Turn Interactions},
  author = {Hu, Yuanzhe and Wang, Yu and McAuley, Julian},
  year = {2025},
  eprint = {2507.05257},
  archivePrefix = {arXiv},
  primaryClass = {cs.CL},
  doi = {10.48550/arXiv.2507.05257}
}

@inproceedings{membench,
  title = {{M}em{B}ench: Towards More Comprehensive Evaluation on the Memory of {LLM}-based Agents},
  author = {Tan, Haoran and Zhang, Zeyu and Ma, Chen and Chen, Xu and Dai, Quanyu and Dong, Zhenhua},
  booktitle = {Findings of the Association for Computational Linguistics: ACL 2025},
  pages = {19336--19352},
  year = {2025},
  publisher = {Association for Computational Linguistics},
  doi = {10.18653/v1/2025.findings-acl.989},
  url = {https://aclanthology.org/2025.findings-acl.989/}
}

@misc{memorybench,
  title = {{MemoryBench}: A Benchmark for Memory and Continual Learning in {LLM} Systems},
  author = {Ai, Qingyao and Tang, Yichen and Wang, Changyue and Long, Jianming and Su, Weihang and Liu, Yiqun},
  year = {2026},
  eprint = {2510.17281},
  archivePrefix = {arXiv},
  primaryClass = {cs.LG},
  doi = {10.48550/arXiv.2510.17281}
}

@misc{mem0,
  title = {Mem0: Building Production-Ready {AI} Agents with Scalable Long-Term Memory},
  author = {Chhikara, Prateek and Khant, Dev and Aryan, Saket and Singh, Taranjeet and Yadav, Deshraj},
  year = {2025},
  eprint = {2504.19413},
  archivePrefix = {arXiv},
  primaryClass = {cs.CL},
  doi = {10.48550/arXiv.2504.19413}
}

@misc{hipporag,
  title = {HippoRAG: Neurobiologically Inspired Long-Term Memory for Large Language Models},
  author = {Jimenez Gutierrez, Bernal and Shu, Yiheng and Gu, Yu and Yasunaga, Michihiro and Su, Yu},
  year = {2024},
  eprint = {2405.14831},
  archivePrefix = {arXiv},
  primaryClass = {cs.CL},
  doi = {10.48550/arXiv.2405.14831}
}

@misc{mirix,
  title = {MIRIX: Multi-Agent Memory System for {LLM}-Based Agents},
  author = {Wang, Yu and Chen, Xi},
  year = {2025},
  eprint = {2507.07957},
  archivePrefix = {arXiv},
  primaryClass = {cs.CL},
  doi = {10.48550/arXiv.2507.07957}
}

@misc{helmet,
  title = {{HELMET}: How to Evaluate Long-Context Language Models Effectively and Thoroughly},
  author = {Yen, Howard and Gao, Tianyu and Hou, Minmin and Ding, Ke and Fleischer, Daniel and Izsak, Peter and Wasserblat, Moshe and Chen, Danqi},
  year = {2025},
  eprint = {2410.02694},
  archivePrefix = {arXiv},
  primaryClass = {cs.CL},
  doi = {10.48550/arXiv.2410.02694},
  note = {ICLR 2025}
}

@misc{memtrack,
  title = {MEMTRACK: Evaluating Long-Term Memory and State Tracking in Multi-Platform Dynamic Agent Environments},
  author = {Deshpande, Darshan and Gangal, Varun and Mehta, Hersh and Kannappan, Anand and Qian, Rebecca and Wang, Peng},
  year = {2025},
  eprint = {2510.01353},
  archivePrefix = {arXiv},
  primaryClass = {cs.AI},
  doi = {10.48550/arXiv.2510.01353},
  note = {NeurIPS 2025 SEA Workshop}
}

@misc{amem,
  title = {A-MEM: Agentic Memory for {LLM} Agents},
  author = {Xu, Wujiang and Liang, Zujie and Mei, Kai and Gao, Hang and Tan, Juntao and Zhang, Yongfeng},
  year = {2025},
  eprint = {2502.12110},
  archivePrefix = {arXiv},
  primaryClass = {cs.CL},
  doi = {10.48550/arXiv.2502.12110},
  note = {NeurIPS 2025}
}

@misc{remem,
  title = {{REMem}: Reasoning with Episodic Memory in Language Agent},
  author = {Shu, Yiheng and Jonnalagedda, Saisri Padmaja and Gao, Xiang and Jim{\'e}nez Guti{\'e}rrez, Bernal and Qi, Weijian and Das, Kamalika and Sun, Huan and Su, Yu},
  year = {2026},
  eprint = {2602.13530},
  archivePrefix = {arXiv},
  primaryClass = {cs.CL},
  note = {ICLR 2026}
}

@misc{simplemem,
  title = {{SimpleMem}: Efficient Lifelong Memory for {LLM} Agents},
  author = {Liu, Jiaqi and Su, Yaofeng and Xia, Peng and Han, Siwei and Zheng, Zeyu and Xie, Cihang and Ding, Mingyu and Yao, Huaxiu},
  year = {2026},
  eprint = {2601.02553},
  archivePrefix = {arXiv},
  primaryClass = {cs.CL}
}

@misc{qwen3,
  title = {{Qwen3} Technical Report},
  author = {{Qwen Team}},
  year = {2025},
  eprint = {2505.09388},
  archivePrefix = {arXiv},
  primaryClass = {cs.CL}
}

@misc{qwen3emb,
  title = {{Qwen3} Embedding: Advancing Text Embedding and Reranking Through Foundation Models},
  author = {{Qwen Team}},
  year = {2025},
  eprint = {2506.05176},
  archivePrefix = {arXiv},
  primaryClass = {cs.CL}
}

@misc{gemini25,
  title = {Gemini 2.5: Pushing the Frontier with Advanced Reasoning, Multimodality, Long Context, and Next Generation Agentic Capabilities},
  author = {{Gemini Team, Google DeepMind}},
  year = {2025},
  eprint = {2507.06261},
  archivePrefix = {arXiv},
  primaryClass = {cs.CL}
}

@article{bm25,
  title = {The Probabilistic Relevance Framework: {BM25} and Beyond},
  author = {Robertson, Stephen and Zaragoza, Hugo},
  journal = {Foundations and Trends in Information Retrieval},
  volume = {3},
  number = {4},
  pages = {333--389},
  year = {2009},
  doi = {10.1561/1500000019}
}

@misc{oai_emb3,
  title = {New Embedding Models and {API} Updates ({text-embedding-3-small})},
  author = {{OpenAI}},
  year = {2024},
  howpublished = {\url{https://openai.com/index/new-embedding-models-and-api-updates/}},
  note = {Model card and API documentation}
}

@misc{oai_gpt5nano,
  title = {Introducing {GPT-5} ({gpt-5-nano} variant)},
  author = {{OpenAI}},
  year = {2025},
  howpublished = {\url{https://openai.com/index/introducing-gpt-5/}},
  note = {Model release announcement and system card}
}

@misc{stale,
  title = {{STALE}: Can {LLM} Agents Know When Their Memories Are No Longer Valid?},
  author = {Chao, Hanxiang and Bai, Yihan and Sheng, Rui and Li, Tianle and Sun, Yushi},
  year = {2026},
  eprint = {2605.06527},
  archivePrefix = {arXiv},
  primaryClass = {cs.CL}
}

@misc{horizonbench,
  title = {{HorizonBench}: Long-Horizon Personalization with Evolving Preferences},
  author = {Li, Shuyue Stella and Paranjape, Bhargavi and Oktar, Kerem and Ma, Zhongyao and Zhou, Gelin and Guan, Lin and Zhang, Na and Park, Sem and Chen, Lin and Yang, Diyi and Tsvetkov, Yulia and Celikyilmaz, Asli},
  year = {2026},
  eprint = {2604.17283},
  archivePrefix = {arXiv},
  primaryClass = {cs.CL}
}

@misc{memorycd,
  title = {{MemoryCD}: Benchmarking Long-Context User Memory of {LLM} Agents for Lifelong Cross-Domain Personalization},
  author = {Zhang, Weizhi and Wei, Xiaokai and Huang, Wei-Chieh and Hui, Zheng and Wang, Chen and Gong, Michelle and Yu, Philip S.},
  year = {2026},
  eprint = {2603.25973},
  archivePrefix = {arXiv},
  primaryClass = {cs.CL}
}

@misc{memt,
  title = {{Mem-T}: Densifying Rewards for Long-Horizon Memory Agents},
  author = {Yue, Yanwei and Zhang, Guibin and Peng, Boci and Fan, Xuanbo and Guo, Jiaxin and Li, Qiankun and Zhang, Yan},
  year = {2026},
  eprint = {2601.23014},
  archivePrefix = {arXiv},
  primaryClass = {cs.LG}
}

@misc{longbench,
  title = {{LongBench}: A Bilingual, Multitask Benchmark for Long Context Understanding},
  author = {Bai, Yushi and Lv, Xin and Zhang, Jiajie and Lyu, Hongchang and Tang, Jiankai and Huang, Zhidian and Du, Zhengxiao and Liu, Xiao and Zeng, Aohan and Hou, Lei and Dong, Yuxiao and Tang, Jie and Li, Juanzi},
  year = {2024},
  eprint = {2308.14508},
  archivePrefix = {arXiv},
  primaryClass = {cs.CL},
  note = {ACL 2024}
}

@misc{infinitebench,
  title = {{InfiniteBench}: Extending Long Context Evaluation Beyond 100K Tokens},
  author = {Zhang, Xinrong and Chen, Yingfa and Hu, Shengding and Xu, Zihang and Chen, Junhao and Hao, Moo and Han, Xu and Thai, Zhen and Wang, Shuo and Liu, Zhiyuan and Sun, Maosong},
  year = {2024},
  eprint = {2402.13718},
  archivePrefix = {arXiv},
  primaryClass = {cs.CL}
}

@misc{ruler,
  title = {{RULER}: What's the Real Context Size of Your Long-Context Language Models?},
  author = {Hsieh, Cheng-Ping and Sun, Simeng and Kriman, Samuel and Acharya, Shantanu and Rekesh, Dima and Jia, Fei and Zhang, Yang and Ginsburg, Boris},
  year = {2024},
  eprint = {2404.06654},
  archivePrefix = {arXiv},
  primaryClass = {cs.CL}
}

@misc{longbenchv2,
  title = {{LongBench v2}: Towards Deeper Understanding and Reasoning on Realistic Long-Context Multitasks},
  author = {Bai, Yushi and Tu, Shangqing and Zhang, Jiajie and Peng, Hao and Cui, Xiaozhi and Wang, Xin and Lv, Xin and Cao, Shulin and Xu, Jiazheng and Liu, Lei and Wang, Zhen and Lv, Chaoyue and Zhang, Yichuan and Liu, Xu and Liu, Xiao and Wang, Yang and Zhang, Ge and Wong, Ka-Hei and Han, Pengcheng and Wang, Chenglei and Chen, Wengyu and Nie, Jian-Yun and Tang, Jie and Li, Juanzi and Hou, Lei and Yuille, Alan and Lin, Shuming},
  year = {2025},
  eprint = {2412.15204},
  archivePrefix = {arXiv},
  primaryClass = {cs.CL},
  note = {ACL 2025}
}

@misc{rag,
  title = {Retrieval-Augmented Generation for Knowledge-Intensive {NLP} Tasks},
  author = {Lewis, Patrick and Perez, Ethan and Piktus, Aleksandra and Petroni, Fabio and Karpukhin, Vladimir and Goyal, Naman and Kuttler, Heinrich and Lewis, Mike and Yih, Wen-tau and Rockt{\"a}schel, Tim and Riedel, Sebastian and Kiela, Douwe},
  year = {2020},
  eprint = {2005.11401},
  archivePrefix = {arXiv},
  primaryClass = {cs.CL},
  note = {NeurIPS 2020}
}

@misc{selfrag,
  title = {Self-{RAG}: Learning to Retrieve, Generate, and Critique through Self-Reflection},
  author = {Asai, Akari and Wu, Zeqiu and Wang, Yizhong and Sil, Avirup and Hajishirzi, Hannaneh},
  year = {2024},
  eprint = {2310.11511},
  archivePrefix = {arXiv},
  primaryClass = {cs.CL},
  note = {ICLR 2024}
}

@misc{generativeagents,
  title = {Generative Agents: Interactive Simulacra of Human Behavior},
  author = {Park, Joon Sung and O'Brien, Joseph C. and Cai, Carrie J. and Morris, Meredith Ringel and Liang, Percy and Bernstein, Michael S.},
  year = {2023},
  eprint = {2304.03442},
  archivePrefix = {arXiv},
  primaryClass = {cs.HC},
  note = {UIST 2023}
}

@misc{memgpt,
  title = {{MemGPT}: Towards {LLM}s as Operating Systems},
  author = {Packer, Charles and Fang, Vivian and Patil, Shishir G. and Lin, Kevin and Wooders, Sarah and Gonzalez, Joseph E.},
  year = {2023},
  eprint = {2310.08560},
  archivePrefix = {arXiv},
  primaryClass = {cs.AI}
}

@misc{memorybank,
  title = {{MemoryBank}: Enhancing Large Language Models with Long-Term Memory},
  author = {Zhong, Wanjun and Guo, Lianghong and Gao, Qiqi and Ye, He and Wang, Yanlin},
  year = {2023},
  eprint = {2305.10250},
  archivePrefix = {arXiv},
  primaryClass = {cs.CL}
}

@misc{realmem,
  title = {{RealMem}: Benchmarking {LLM}s in Real-World Memory-Driven Interaction},
  author = {Bian, Haonan and Yao, Zhiyuan and Hu, Sen and Xu, Zishan and Zhang, Shaolei and Guo, Yifu and Yang, Ziliang and Han, Xueran and Wang, Huacan and Chen, Ronghao},
  year = {2026},
  eprint = {2601.06966},
  archivePrefix = {arXiv},
  primaryClass = {cs.CL}
}

@misc{mobilemem,
  title = {{MobileMem}: Evaluating Long-Horizon Memory for Language Agents in Real-World Mobile Environments},
  author = {Deng, Xinle and Xue, Yida and Chen, Yijun and Mao, Mingjun and Zhong, Ruobin and Xu, Buqiang and Fang, Jizhan and Xu, Haoming and Wu, Tingwei and Xu, Yajing and Deng, Shumin and Wang, Haofen and Chen, Huajun and Zhang, Ningyu},
  year = {2026},
  howpublished = {OpenReview},
  url = {https://openreview.net/forum?id=w5I11HrMgJ},
  note = {ICLR 2026 Lifelong Agent Workshop}
}

@inproceedings{msc,
  title = {Beyond Goldfish Memory: Long-Term Open-Domain Conversation},
  author = {Xu, Jing and Szlam, Arthur and Weston, Jason},
  booktitle = {Proceedings of the 60th Annual Meeting of the Association for Computational Linguistics (Volume 1: Long Papers)},
  pages = {5180--5197},
  year = {2022},
  publisher = {Association for Computational Linguistics},
  doi = {10.18653/v1/2022.acl-long.356},
  url = {https://aclanthology.org/2022.acl-long.356/}
}

@misc{esmemeval,
  title = {{ES-MemEval}: Benchmarking Conversational Agents on Personalized Long-Term Emotional Support},
  author = {Chen, Tiantian and Lu, Jiaqi and Shen, Ying and Zhang, Lin},
  year = {2026},
  eprint = {2602.01885},
  archivePrefix = {arXiv},
  primaryClass = {cs.CL}
}

\section{Use of AI Writing Assistance}
\label{app:ai-assistance}
Consistent with the policy on AI writing assistance, we used AI assistants only
for code development and for polishing the wording and grammar of the
manuscript. All scientific content---the claims, experimental design, results,
and analyses---was conceived, written, reviewed, and verified by the human
authors, who take full responsibility for it.

\appendix
\raggedbottom
\section{Benchmark Details}
\label{app:benchmark-details}

The final benchmark index listed in the artifact manifest contains 20 users, 835
knowledge points, 5,677 base probes, and 15,422 question rows. Paper-facing
loaders use the scoring protocol listed in the same manifest.

\subsection{Scale and Workload}

This subsection reports the benchmark scale and scoring workload referenced
from \Cref{sec:benchmark-data}.

\begin{table}[H]
\centering
\scriptsize
\setlength{\tabcolsep}{3pt}
\renewcommand{\arraystretch}{1.02}
\begin{tabular*}{\columnwidth}{@{\extracolsep{\fill}}lrlr@{}}
\toprule
\multicolumn{2}{@{}l}{\textit{Scale}} &
\multicolumn{2}{l@{}}{\textit{Workload}} \\
\midrule
Users & 20 & Main cfgs. & 13 \\
KPs & 835 & Paradigms & 4 \\
Base probes & 5,677 & Scored answers & 200,453 \\
Question rows & 15,422 & Windows/user & 8 \\
\bottomrule
\end{tabular*}
\caption{Benchmark scale and scoring workload. Base probes are
knowledge-point/window probes; question rows are produced after expanding
probes by question type and evidence condition. Scored system-answer rows sum
the completed 13 main configurations under the paper-facing scoring pass.}
\label{tab:benchmark-composition}
\end{table}

\subsection{Quality-Control Pipeline}

\benchmark treats quality control as a construction check rather than a result
metric. The pipeline first preserves grounded source sessions and source memory
evidence, then transforms them into typed knowledge points and repeated
window-level probes. Automated checks flag candidate grounding, leakage,
temporal-anchor, type, and semantic-consistency defects; flagged cases are
manually reviewed and corrected before the benchmark index is frozen. The check
summary is reported below, and the independent second-judge scoring check is
reported in Appendix~D.

\begin{table}[H]
\centering
\small
\resizebox{\columnwidth}{!}{
\begin{tabular}{lrrr}
\toprule
Audit dimension & Threshold & Pre-fix fail & Post-fix fail \\
\midrule
Grounding                       & $<5\%$  & 0.16\%  & \textbf{0.00\%} \\
Answer leakage (token)          & $<10\%$ & 6.76\%  & \textbf{2.79\%} \\
Answer leakage (semantic)       & $<10\%$ & --      & \textbf{0.00\%} \\
Cross-question differentiation  & $<5\%$  & 0.00\%  & \textbf{0.00\%} \\
Temporal anchor                 & $<15\%$ & 38.74\% & \textbf{1.42\%} \\
Type correctness                & $<2\%$  & 0.00\%  & \textbf{0.00\%} \\
Semantic consistency            & $<5\%$  & 10.24\% & \textbf{0.00\%} \\
Format completeness             & $0\%$   & 0.00\%  & \textbf{0.00\%} \\
\bottomrule
\end{tabular}
}
\caption{Quality-control check applied during \benchmark construction. Pre-fix
shows defects inherited from the HaluMem-Medium-derived draft; Post-fix is after
manual review and correction.}
\label{tab:fidelity}
\end{table}

\FloatBarrier

\section{System Configurations}
\label{app:system-configurations}

\Cref{tab:systems} lists the 13 main memory-system configurations used in the
controlled comparison. The comparison unit is a complete configuration: memory
construction, retrieval or access path, update policy, answer prompt, and
answer generator. Long-context rows are provider-native; the remaining main
rows use the shared gpt-4o-mini answer generator described in
\Cref{sec:experiment-setting}.

\begin{table*}[t]
\centering
\scriptsize
\resizebox{\textwidth}{!}{
\begin{tabular}{@{}p{0.11\linewidth}p{0.11\linewidth}p{0.10\linewidth}p{0.13\linewidth}p{0.13\linewidth}p{0.12\linewidth}p{0.09\linewidth}p{0.09\linewidth}@{}}
\toprule
System & Paradigm & Backbone & Memory unit & Retrieval & Update mech. & Conflict path & Temporal path \\
\midrule
Qwen3.5-35B            & Long-Context     & native    & full context           & --                    & --                    & partial & partial \\
Gemini-3-Flash              & Long-Context     & native    & full context           & --                    & --                    & partial & partial \\
GPT-5-nano          & Long-Context     & native    & full context           & --                    & --                    & partial & partial \\
\midrule
BM25                & RAG              & 4o-mini   & chunk                  & sparse (BM25)         & append                & no      & no      \\
Text-emb-3-small    & RAG              & 4o-mini   & chunk                  & dense                 & append                & no      & no      \\
Qwen3-Emb           & RAG              & 4o-mini   & chunk                  & dense                 & append                & no      & no      \\
HippoRAG-v2            & RAG              & 4o-mini   & event graph (KG)       & hybrid (graph+dense)  & append                & partial & no      \\
\midrule
Mem0                & External-Memory  & 4o-mini   & fact triple            & dense                 & consolidate           & partial & no      \\
SimpleMem           & External-Memory  & 4o-mini   & compressed note        & dense                 & consolidate           & partial & partial \\
REMem               & External-Memory  & 4o-mini   & event graph (gist+fact)& hybrid (agentic)      & consolidate           & yes     & yes     \\
AMem                & External-Memory  & 4o-mini   & note                   & dense (Chroma)        & append                & partial & no      \\
\midrule
MIRIX               & Agentic-Memory   & 4o-mini   & 6-store scratchpad     & hybrid                & consolidate           & partial & partial \\
Mem-T               & Agentic-Memory   & 4o-mini   & note (Chroma)          & dense (RL policy)     & consolidate           & partial & partial \\
\bottomrule
\end{tabular}
}
\caption{System-design matrix for the 13 evaluated main configurations.
Backbone disclosures are used when interpreting cross-system comparisons.
Conflict path and temporal path are descriptive implementation labels: ``yes''
denotes an explicit mechanism or prompt path for that axis, ``partial'' denotes
an indirect or general mechanism, and ``no'' denotes no dedicated
axis-specific mechanism in our implementation. They are not standalone
capability claims.}
\label{tab:systems}
\end{table*}

Mem-T is integrated from the released Mem-T-4B checkpoint with the published
MOT-GRPO retrieval policy and hindsight-credit-assignment memory construction.
Its hierarchical memory contains working, factual, experiential, and raw stores,
all backed by ChromaDB. Mem-T's published in-domain training corpus is LoCoMo;
\benchmark is derived from HaluMem-Medium with LoCoMo session lineage, so the
released policy may have soft distribution overlap with our evaluation data. We
use the frozen checkpoint and do not fine-tune on \benchmark probes.

\section{Full Results and Failure-Origin Checks}
\label{app:full-results}

This appendix provides full numeric companions to the compact main-text tables
and figures. Values use the main MemTrace adapter configurations unless stated
otherwise. It also includes supporting views for endpoints, hallucination,
conflict-by-window behavior, abstention precision, and qualitative examples.

\subsection{Full Memory-Age and Evidence-Condition Results}

These tables expand the compact main-text results without changing the
aggregation convention. \Cref{tab:appendix-full-leaderboard} reports the full
Fresh/Saturated/\deltaforget profile, \Cref{tab:appendix-conflict-boundary-full}
adds confidence intervals for conflict and boundary probes, and
\Cref{tab:appendix-rank-robustness} shows the rank stability behind the
question-type comparison.

\begin{table*}[t]
\centering
\setlength{\tabcolsep}{4pt}
\renewcommand{\arraystretch}{0.94}
\begin{tabular}{llrrrr}
\toprule
System & Paradigm & Question type & Fresh & Saturated & $\Delta$Forget \\
\midrule
HippoRAG-v2 & RAG & Current & 69.7 & 45.4 & 24.4 \\
 &  & Historical & 67.6 & 50.9 & 16.7 \\
 &  & Trajectory & 33.0 & 13.4 & 19.7 \\
\midrule
Mem-T & Agentic Memory & Current & 58.9 & 40.4 & 18.5 \\
 &  & Historical & 59.8 & 47.3 & 12.5 \\
 &  & Trajectory & 47.2 & 19.8 & 27.4 \\
\midrule
SimpleMem & External Memory & Current & 60.4 & 43.8 & 16.6 \\
 &  & Historical & 58.0 & 44.8 & 13.2 \\
 &  & Trajectory & 40.1 & 17.3 & 22.8 \\
\midrule
text-emb-3-small & RAG & Current & 65.7 & 43.4 & 22.3 \\
 &  & Historical & 59.3 & 47.5 & 11.8 \\
 &  & Trajectory & 19.7 & 9.4 & 10.3 \\
\midrule
REMem & External Memory & Current & 62.1 & 43.1 & 18.9 \\
 &  & Historical & 53.7 & 43.1 & 10.6 \\
 &  & Trajectory & 24.1 & 12.6 & 11.5 \\
\midrule
Qwen3-Emb & RAG & Current & 61.2 & 41.1 & 20.1 \\
 &  & Historical & 59.2 & 48.2 & 11.0 \\
 &  & Trajectory & 15.7 & 8.6 & 7.1 \\
\midrule
Gemini-3-Flash & Long Context & Current & 66.0 & 39.7 & 26.2 \\
 &  & Historical & 59.6 & 47.1 & 12.5 \\
 &  & Trajectory & 31.5 & 11.0 & 20.5 \\
\midrule
BM25 & RAG & Current & 61.2 & 39.9 & 21.3 \\
 &  & Historical & 56.2 & 46.9 & 9.3 \\
 &  & Trajectory & 18.3 & 9.2 & 9.1 \\
\midrule
AMem & External Memory & Current & 55.3 & 42.0 & 13.3 \\
 &  & Historical & 47.7 & 39.3 & 8.5 \\
 &  & Trajectory & 15.6 & 10.9 & 4.7 \\
\midrule
Qwen3.5-35B & Long Context & Current & 68.8 & 32.7 & 36.1 \\
 &  & Historical & 66.2 & 44.2 & 21.9 \\
 &  & Trajectory & 49.0 & 6.7 & 42.3 \\
\midrule
GPT-5-nano & Long Context & Current & 67.8 & 34.8 & 33.0 \\
 &  & Historical & 62.4 & 41.9 & 20.5 \\
 &  & Trajectory & 38.4 & 6.5 & 31.9 \\
\midrule
Mem0 & External Memory & Current & 50.1 & 31.5 & 18.7 \\
 &  & Historical & 37.5 & 29.7 & 7.7 \\
 &  & Trajectory & 9.6 & 8.2 & 1.5 \\
\midrule
MIRIX & Agentic Memory & Current & 56.4 & 26.4 & 30.0 \\
 &  & Historical & 40.5 & 23.8 & 16.7 \\
 &  & Trajectory & 12.4 & 4.8 & 7.6 \\
\bottomrule
\end{tabular}

\caption{Full main-adapter Fresh, Saturated, and \deltaforget table by question
type. Fresh/Saturated Gist is shown as percentages and \deltaforget as
percentage points. Systems are ordered by mean Saturated Gist in the main
leaderboard.}
\label{tab:appendix-full-leaderboard}
\end{table*}

\begin{table*}[t]
\centering
\footnotesize
\resizebox{\textwidth}{!}{
\setlength{\tabcolsep}{3pt}
\begin{tabular}{lcccccc}
\toprule
 & \multicolumn{3}{c}{Conflict (W7--W8)} & \multicolumn{3}{c}{Boundary} \\
\cmidrule(lr){2-4} \cmidrule(lr){5-7}
System & Gist & Abst & Hallu & Gist & Abst & Hallu \\
\midrule
Qwen3.5-35B & 88.6\,[84.9,92.2] & 8.4\,[6.4,10.6] & 3.9\,[1.9,5.9] & 89.6\,[79.6,97.8] & 24.8\,[18.9,30.8] & 10.4\,[2.3,21.3] \\
HippoRAG-v2 & 80.2\,[73.9,85.9] & 11.7\,[8.1,15.7] & 11.1\,[6.6,15.9] & 94.4\,[88.9,98.4] & 57.5\,[50.1,64.0] & 5.2\,[1.1,10.0] \\
Gemini-3-Flash & 79.8\,[75.2,84.0] & 15.2\,[12.4,18.6] & 6.4\,[3.2,9.9] & 96.0\,[92.6,98.7] & 90.3\,[85.9,94.0] & 3.8\,[1.1,7.1] \\
GPT-5-nano & 79.0\,[73.9,83.3] & 16.9\,[13.6,19.9] & 5.7\,[3.5,8.1] & 95.1\,[91.2,98.1] & 68.8\,[64.2,73.1] & 4.9\,[1.9,9.1] \\
BM25 & 76.7\,[72.4,80.6] & 20.3\,[15.9,24.8] & 3.6\,[1.9,5.5] & 96.6\,[92.2,99.5] & 96.4\,[92.4,99.4] & 3.4\,[0.6,7.5] \\
SimpleMem & 74.0\,[67.9,80.3] & 13.9\,[9.0,19.2] & 12.6\,[8.6,17.1] & 91.6\,[86.4,95.6] & 28.7\,[21.2,35.9] & 7.6\,[3.1,12.6] \\
Mem-T & 73.0\,[67.9,78.1] & 18.3\,[15.6,21.2] & 11.4\,[8.0,15.1] & 91.1\,[87.5,94.1] & 69.7\,[65.8,73.8] & 8.6\,[5.4,12.8] \\
Qwen3-Emb & 69.8\,[64.5,74.6] & 23.1\,[18.2,29.1] & 8.0\,[5.1,11.1] & 95.8\,[91.7,99.4] & 94.5\,[89.7,98.1] & 4.2\,[0.8,9.2] \\
Text-emb-3-small & 68.3\,[62.6,73.8] & 25.9\,[20.8,31.5] & 6.9\,[3.6,10.6] & 96.4\,[92.1,99.6] & 95.6\,[91.1,98.9] & 3.6\,[0.4,8.0] \\
MIRIX & 57.1\,[51.7,62.4] & 27.5\,[22.4,32.6] & 18.1\,[13.5,23.1] & 97.1\,[94.4,99.2] & 96.9\,[94.4,99.1] & 2.9\,[0.8,5.6] \\
REMem & 35.1\,[28.5,42.4] & 61.8\,[55.0,68.4] & 4.9\,[2.1,8.1] & 94.5\,[88.0,98.9] & 94.0\,[90.4,97.2] & 3.6\,[0.9,6.5] \\
AMem & 20.1\,[12.6,27.4] & 78.3\,[70.6,85.7] & 1.1\,[0.2,2.1] & 97.4\,[93.8,99.9] & 97.4\,[93.4,99.9] & 2.6\,[0.1,6.3] \\
Mem0 & 14.6\,[10.1,19.1] & 82.5\,[77.8,87.9] & 2.7\,[0.8,4.8] & 99.3\,[97.9,100.0] & 99.3\,[97.9,100.0] & 0.7\,[0.0,2.1] \\
\bottomrule
\end{tabular}
}
\caption{Main-adapter 13-system conflict and boundary table with bootstrap
95\% confidence intervals. Cells report percentage [lower, upper]. This is the
confidence-interval companion to \Cref{tab:conflict-boundary-profile}.}
\label{tab:appendix-conflict-boundary-full}
\end{table*}

\begin{table*}[t]
\centering
\footnotesize
\resizebox{\textwidth}{!}{
\setlength{\tabcolsep}{4pt}
\begin{tabular}{llccc}
\toprule
System & Paradigm & Current Sat. [95\% CI] & Historical Sat. [95\% CI] & Trajectory Sat. [95\% CI] \\
\midrule
HippoRAG-v2 & RAG & 45.4\,[40.5,50.2] & 50.9\,[46.4,55.3] & 13.4\,[10.4,16.3] \\
Mem-T & Agentic Memory & 40.4\,[35.3,45.5] & 47.3\,[43.8,50.7] & 19.8\,[16.3,23.3] \\
SimpleMem & External Memory & 43.8\,[37.9,49.7] & 44.8\,[37.9,51.7] & 17.3\,[13.9,20.8] \\
text-emb-3-small & RAG & 43.4\,[39.6,47.2] & 47.5\,[42.3,52.6] & 9.4\,[7.0,11.8] \\
REMem & External Memory & 43.1\,[39.2,47.1] & 43.1\,[40.0,46.2] & 12.6\,[9.7,15.6] \\
Qwen3-Emb & RAG & 41.1\,[35.9,46.2] & 48.2\,[43.9,52.5] & 8.6\,[6.5,10.7] \\
Gemini-3-Flash & Long Context & 39.7\,[36.4,43.0] & 47.1\,[43.7,50.6] & 11.0\,[8.7,13.3] \\
BM25 & RAG & 39.9\,[36.0,43.9] & 46.9\,[41.7,52.0] & 9.2\,[6.8,11.6] \\
AMem & External Memory & 42.0\,[36.6,47.4] & 39.3\,[35.0,43.6] & 10.9\,[7.9,13.8] \\
Qwen3.5-35B & Long Context & 32.7\,[28.4,36.9] & 44.2\,[40.0,48.5] & 6.7\,[4.7,8.7] \\
GPT-5-nano & Long Context & 34.8\,[30.9,38.7] & 41.9\,[37.0,46.8] & 6.5\,[4.5,8.5] \\
Mem0 & External Memory & 31.5\,[27.8,35.1] & 29.7\,[25.4,34.1] & 8.2\,[6.3,10.0] \\
MIRIX & Agentic Memory & 26.4\,[23.5,29.4] & 23.8\,[20.3,27.4] & 4.8\,[3.2,6.4] \\
\midrule
\multicolumn{5}{l}{\textit{Rank-correlation across question types (1{,}000-fold user bootstrap; from followup\_robustness\_report.md \S 11.4)}} \\
\multicolumn{2}{l}{Comparison} & \multicolumn{1}{c}{Median $\rho$} & \multicolumn{1}{c}{95\% CI} & \multicolumn{1}{c}{\% boots $< 0.6$} \\
\multicolumn{2}{l}{Cur+Hist mean vs Traj} & 0.473 & [0.247, 0.747] & 82.1\% \\
\multicolumn{2}{l}{Current vs Trajectory} & 0.516 & [0.291, 0.736] & 77.8\% \\
\multicolumn{2}{l}{Historical vs Trajectory} & 0.385 & [0.192, 0.692] & 91.1\% \\
\bottomrule
\end{tabular}
}
\caption{Per-system bootstrap 95\% confidence intervals on main-adapter
\saturated Gist by question type, shown as percentages, and rank-correlation
distribution between current/historical and trajectory ranks.}
\label{tab:appendix-rank-robustness}
\end{table*}

\subsection{Additional Supporting Views}

\begin{figure*}[t]
\centering
\includegraphics[width=\textwidth]{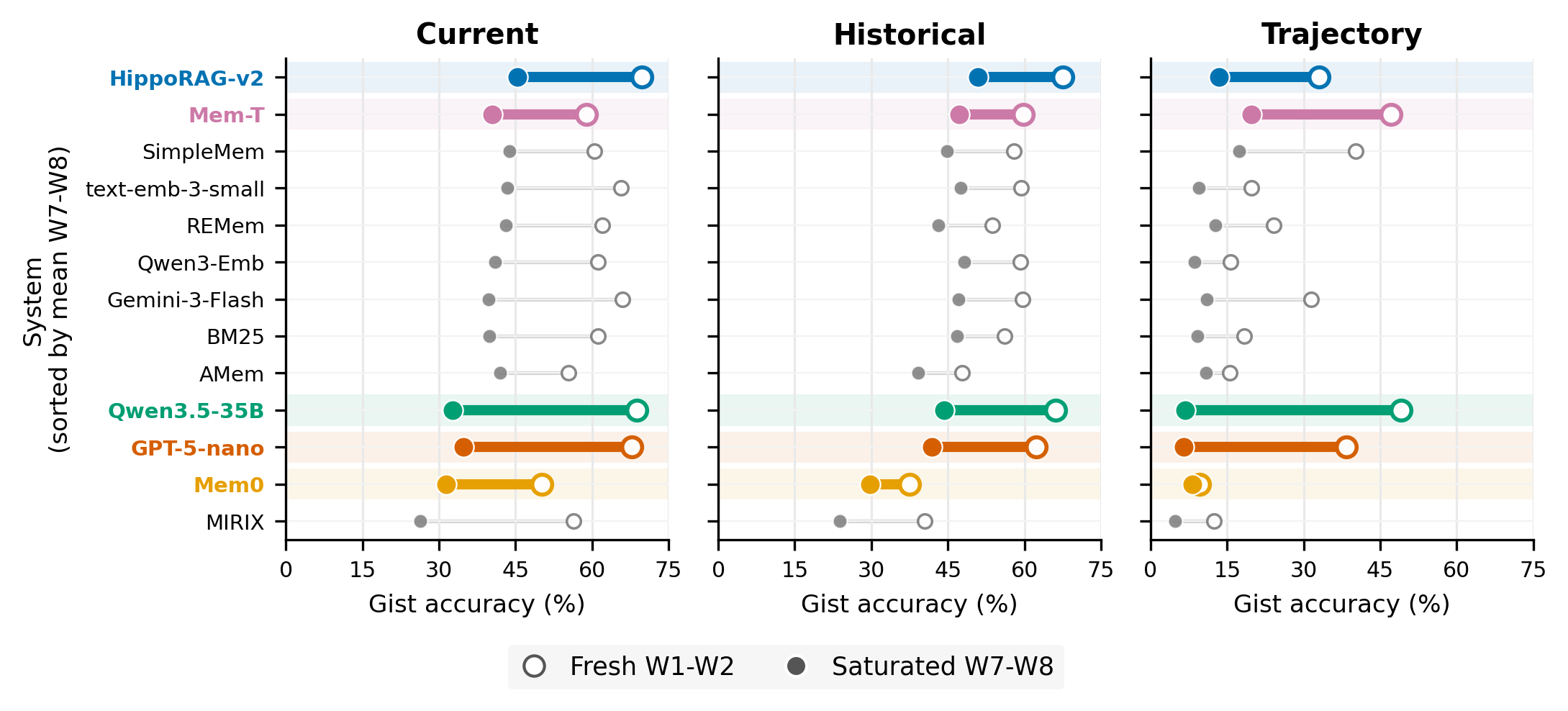}
\caption{Per-system \fresh-to-\saturated endpoints by question type for the
main \benchmark adapter configurations. Filled markers denote \fresh W1--W2
and hollow markers denote \saturated W7--W8. This is the endpoint-level
companion to the main memory-age trace in \Cref{fig:window-retention}.}
\label{fig:appendix-fresh-sat-endpoints}
\end{figure*}

\begin{figure*}[t]
\centering
\includegraphics[width=0.9\textwidth]{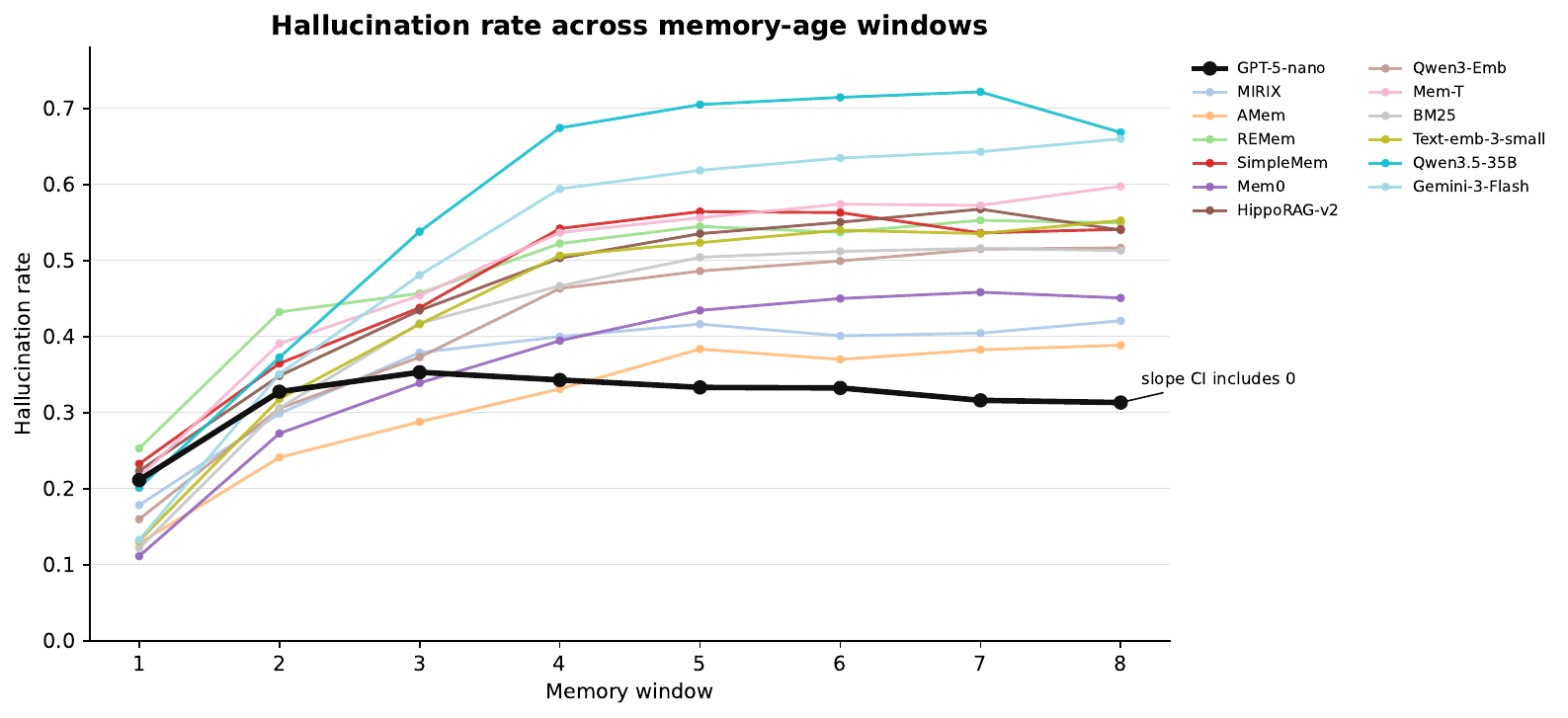}
\vspace{2pt}

\includegraphics[width=0.9\textwidth]{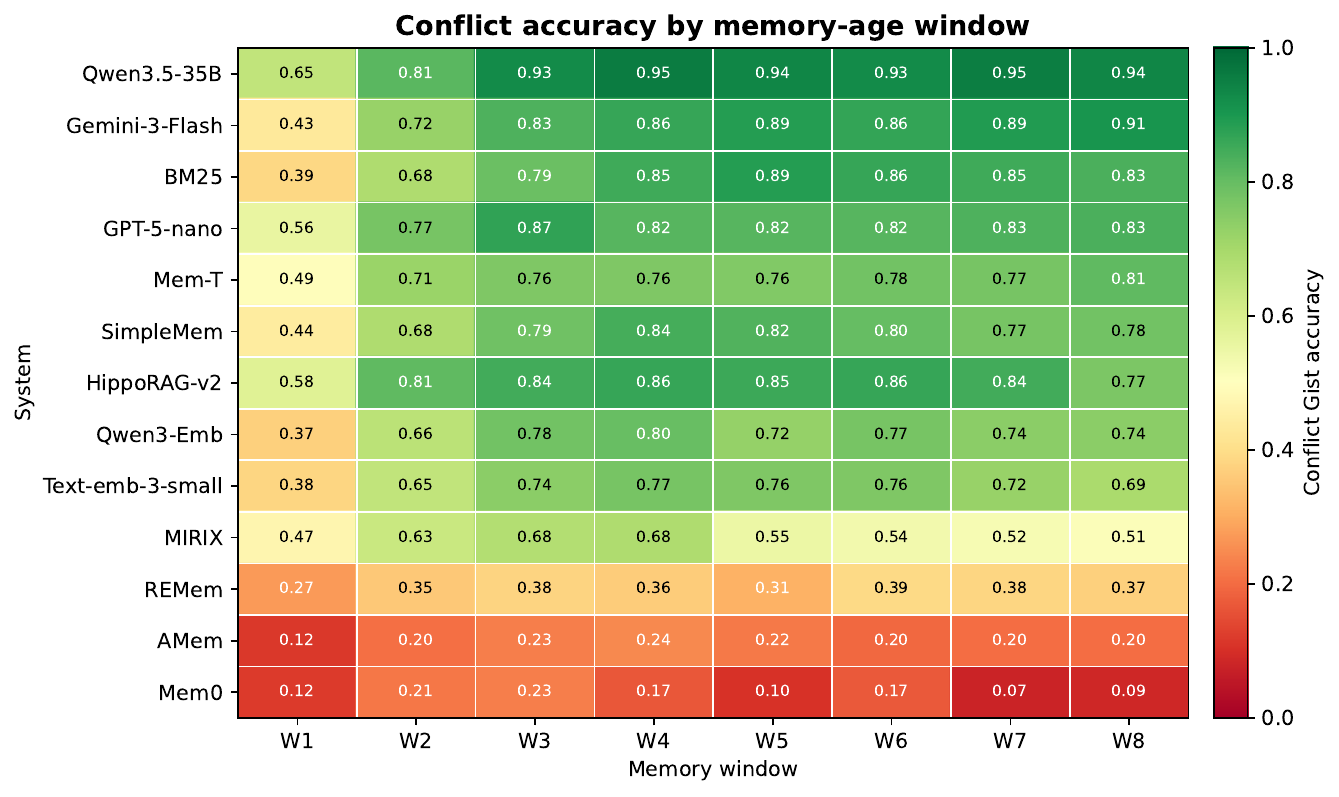}
\caption{Additional memory-window diagnostics. Top: main-adapter per-system
hallucination rate across W1--W8. Bottom: main-adapter Conflict Gist accuracy
by system and memory window, expanding the conflict/boundary profile and its
confidence-interval companion table.}
\label{fig:appendix-hallu-by-window}
\label{fig:appendix-age-conflict}
\end{figure*}

The following figures and tables preserve supporting results that are too large
for the main text. \Cref{fig:appendix-fresh-sat-endpoints} shows the
Fresh-to-Saturated endpoints by question type,
\Cref{fig:appendix-hallu-by-window,fig:appendix-age-conflict} shows
additional memory-window diagnostics,
\Cref{tab:appendix-abstention-precision} reports abstention precision, and
\Cref{tab:appendix-showcases} gives representative qualitative probes.

\subsection{Failure-Origin Details}

The failure-origin analysis has two parts. The oracle block asks whether the
answer generator can recover when the relevant memory evidence is supplied
directly. The replay block then checks, with a transparent Text-emb-3-small
proxy, whether the same evidence was reachable in the original production
setting.

\begin{table}[H]
\centering
\scriptsize
\setlength{\tabcolsep}{3pt}
\resizebox{\columnwidth}{!}{%
\begin{tabular}{@{}lrrrr@{}}
\toprule
Oracle subset & $n$ & Base & Oracle & Lift \\
\midrule
Trajectory overall & 80 & 16.9\% & 82.5\% & +65.6 \\
All-13-fail traj. & 40 & 0.0\% & 85.0\% & +85.0 \\
Dynamic-changed traj. & 40 & 33.8\% & 80.0\% & +46.2 \\
Hard cur./hist. & 40 & 8.5\% & 80.0\% & +71.5 \\
\midrule
\multicolumn{5}{@{}l}{\textit{Reach/use replay with Text-emb-3-small}} \\
Evidence reach & 300 & -- & 93.0\% & -- \\
Reach gap ($R=0$) & 300 & -- & 7.0\% & [3.8, 10.6] \\
Use gap ($R=1,U=0$) & 300 & -- & 73.3\% & [67.1, 79.2] \\
Success ($R=1,U=1$) & 300 & -- & 19.7\% & [14.8, 25.3] \\
Use-gap range & -- & -- & 69--88\% & -- \\
\bottomrule
\end{tabular}
}
\caption{Numerical companion to the main oracle/recovery decomposition figure.
\emph{Base} is the mean production-system solve rate; \emph{Oracle} is Gist
accuracy when relevant memory evidence is supplied directly. The replay rows
use the transparent Text-emb-3-small reach proxy: $R$ marks whether the proxy
reaches the benchmark source session for the gold answer, and $U$ marks whether
the original production answer solves the probe. Lift and range values are
percentage points.}
\label{tab:appendix-oracle-recovery-trace}
\end{table}

\section{Judge Reliability}
\label{app:judge-reliability}

\benchmark uses GPT-4o as the primary scoring judge. As a reliability check, we
rescore a stratified 200-probe sample with a second LLM judge using the same
rubric and JSON schema, then compute Cohen's $\kappa$ with bootstrap confidence
intervals.

\begin{table}[hp]
\centering
\small
\resizebox{\columnwidth}{!}{
\begin{tabular}{@{}lrrc@{}}
\toprule
Dimension & $n$ & Cohen's $\kappa$ & 95\% bootstrap CI \\
\midrule
Gist accuracy & 200 & 0.772 & [0.680, 0.858] \\
Response type & 200 & 0.703 & [0.623, 0.785] \\
\bottomrule
\end{tabular}
}
\caption{Second-judge reliability on a stratified 200-probe sample. Agreement is
computed between the primary GPT-4o judge and a Gemini-3-Flash cross-judge using the
same scoring rubric.}
\label{tab:judge-reliability}
\end{table}

\begin{table}[t]
\centering
\scriptsize
\setlength{\tabcolsep}{8pt}
\begin{tabular}{llrrr}
\toprule
Question type & KP type & $n$ & Gist disag. & Any disag. \\
\midrule
Trajectory & Dynamic & 24 & 6 & 45.8\% \\
Current & Static & 23 & 5 & 43.5\% \\
Historical & Static & 21 & 3 & 33.3\% \\
Current & Preference & 22 & 3 & 27.3\% \\
Historical & Preference & 22 & 4 & 27.3\% \\
Historical & Dynamic & 24 & 1 & 25.0\% \\
Current & Dynamic & 22 & 0 & 18.2\% \\
Trajectory & Preference & 19 & 0 & 5.3\% \\
Trajectory & Static & 23 & 0 & 4.3\% \\
\bottomrule
\end{tabular}
\caption{Second-judge disagreement breakdown on the same 200-probe sample.
``Any disagreement'' counts probes where the judges disagree on either binary
Gist accuracy or response type.}
\label{tab:judge-disagreement-breakdown}
\end{table}

\section{Sensitivity Checks}
\label{app:sensitivity-checks}

The main paper reports controlled MemTrace adapter configurations. The checks
below vary prompts, paper-native settings, answer backbones, or diagnostic
setup to quantify sensitivity around those main rows. They are not counted as
additional benchmark systems.

\subsection{Configuration and Prompt Sensitivity}
\label{app:configuration-checks}

This section keeps the main benchmark rows fixed and reports variants that
stress prompts, paper-native settings, and matched results. The goal is to make
clear which conclusions are stable and which numbers should be read only as
configuration checks rather than main-system replacements.

\subsection{Backbone and Oracle Sensitivity}
\label{app:backbone-sensitivity}

We also check whether the conflict and oracle conclusions depend on the answer
backbone. These checks swap only the answer-time generator while keeping the
rest of the diagnostic setup fixed.

\subsection{Evidence-Interface and Actionability Checks}
\label{app:interface-actionability}

The following checks are diagnostic interventions, not additional benchmark
systems. They ask whether changing the evidence interface or using oracle
benchmark labels can change the observed failure pattern.

\begin{table*}[t]
\centering
\scriptsize
\resizebox{\textwidth}{!}{%
\begin{tabular}{@{}lllll@{}}
\toprule
System & Main adapter & Variant & Coverage & Reading \\
\midrule
HippoRAG-v2 & main 42.4\% & unified 38.3\% & 20 users & Unified prompt lowers overall Gist. \\
REMem & main 36.3\%; paper config 36.7\% & unified 29.9\% & 20 users & Paper config ties main; unified is more conservative. \\
SimpleMem & main 39.0\% & unified 26.7\%; gpt-4.1-mini 37.1\% & 20 users & Prompt/backbone sensitive. \\
MIRIX & main 22.5\% & gpt-4.1-mini 37.4\% & 20 users & Paper-native backbone is stronger. \\
Mem-T & main 40.4\%; matched subset 41.0\% & unified 42.1\% & 20 users & Paper config \\
\bottomrule
\end{tabular}}
\caption{Configuration sensitivity checks. Values are overall Gist under the
core-KP user-macro mean-over-question-types convention from the ablation
inventory, shown as percentages. These variants are not mixed into the main
leaderboard.}
\label{tab:appendix-config-checks}
\end{table*}

\begin{table*}[t]
\centering
\scriptsize
\resizebox{\textwidth}{!}{%
\begin{tabular}{@{}llllll@{}}
\toprule
System & Conflict Gist & Conflict Abst. & Boundary Abst. & Coverage & Reading \\
\midrule
HippoRAG-v2 & 80.2\% $\rightarrow$ 33.8\% & 11.7\% $\rightarrow$ 62.7\% & 57.5\% $\rightarrow$ 96.5\% & 20 users & Native is more conflict-committing. \\
REMem & 35.1\% $\rightarrow$ 13.6\% & 61.8\% $\rightarrow$ 83.2\% & 94.0\% $\rightarrow$ 99.9\% & 20 users & Unified is more abstention-heavy. \\
SimpleMem & 74.0\% $\rightarrow$ 63.4\% & 13.9\% $\rightarrow$ 33.7\% & 28.7\% $\rightarrow$ 95.9\% & 20 users & Unified trades conflict commitment for refusal. \\
Mem-T & 72.0\% $\rightarrow$ 64.7\% & 20.0\% $\rightarrow$ 32.0\% & 67.0\% $\rightarrow$ 92.7\% & 20 users &  safer but less committing. \\
\bottomrule
\end{tabular}}
\caption{Distractor prompt-control checks. Arrows report main adapter
$\rightarrow$ unified where available. The table shows configuration
sensitivity for the conflict/boundary results, with rates shown as
percentages.}
\label{tab:appendix-distractor-config-checks}
\end{table*}

\begin{figure*}[t]
\centering
\includegraphics[width=\textwidth]{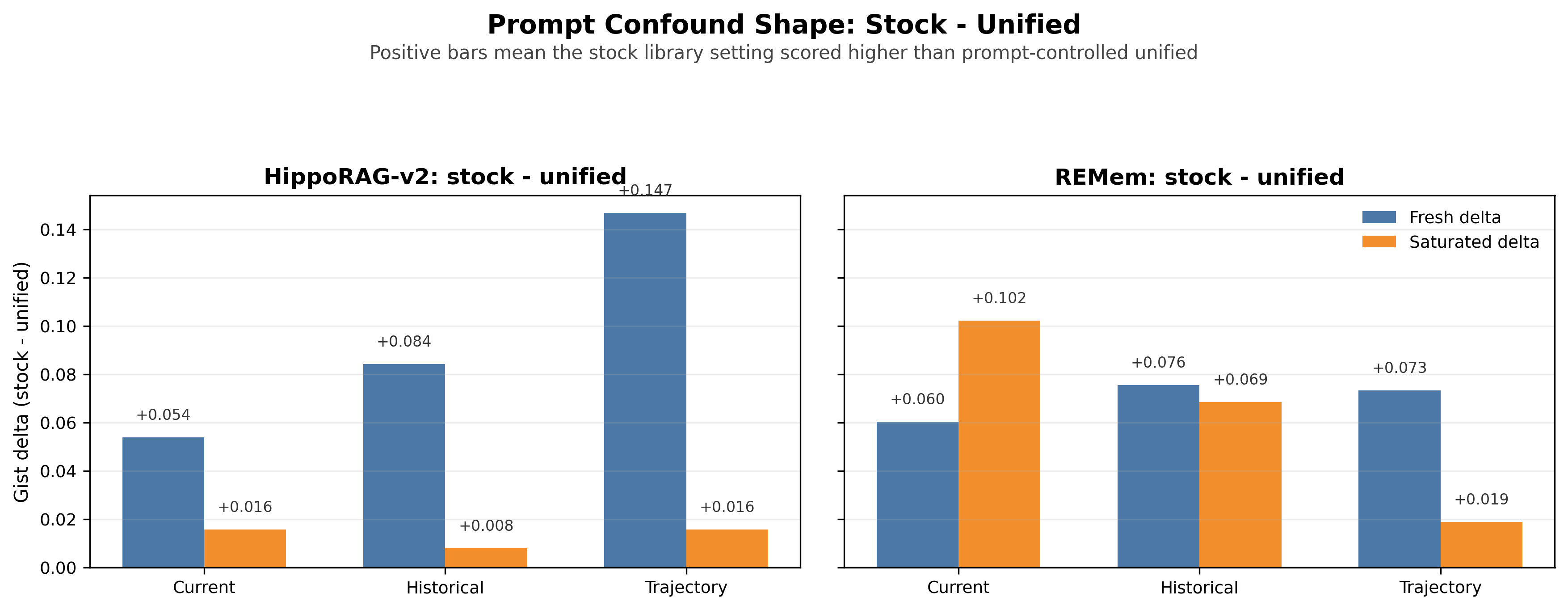}
\caption{Prompt-control ablation by horizon and question type. This sensitivity
check asks whether horizon and question-form changes alone explain the
retention patterns; it does not replace the main adapter Fresh-to-Saturated
trace.}
\label{fig:appendix-prompt-confound}
\end{figure*}

\begin{table*}[t]
\centering
\scriptsize
\resizebox{\textwidth}{!}{%
\begin{tabular}{llrccc}
\toprule
Scope & Variant & $n$ & Gist [95\% CI] & Lift [95\% CI] & Secondary change \\
\midrule
Saturated trajectory & baseline & 426 & 20.0 [14.2, 26.2] & -- & Hallu. 71.3; Abst. 8.9 \\
Saturated trajectory & tr+cr+ac & 426 & 31.2 [22.4, 39.9] & +11.1 [+4.4, +18.3] & Hallu. -5.8; Abst. -6.1 \\
Conflict & baseline & 800 & 68.1 [62.0, 73.9] & -- & Hallu. 7.0; Abst. 25.9 \\
Conflict & cr & 800 & 73.6 [68.6, 78.2] & +5.5 [+1.4, +9.9] & Hallu. -2.2; Abst. +0.7 \\
Boundary & baseline & 800 & 96.6 [92.7, 99.6] & -- & Hallu. 3.4; Abst. 95.8 \\
Boundary & tr & 800 & 97.3 [93.4, 100.0] & +0.6 [+0.0, +1.5] & Hallu. -0.6; Abst. +1.1 \\
\bottomrule
\end{tabular}}
\caption{Diagnostic actionability check using benchmark labels as oracle
routing signals on top of the transparent dense-RAG baseline. Variants are not
deployable memory methods and are not counted as benchmarked systems; they
only test whether the diagnostic axes can guide targeted interventions.}
\label{tab:appendix-plugin-actionability}
\end{table*}

\begin{figure*}[htbp]
\centering
\includegraphics[width=\textwidth]{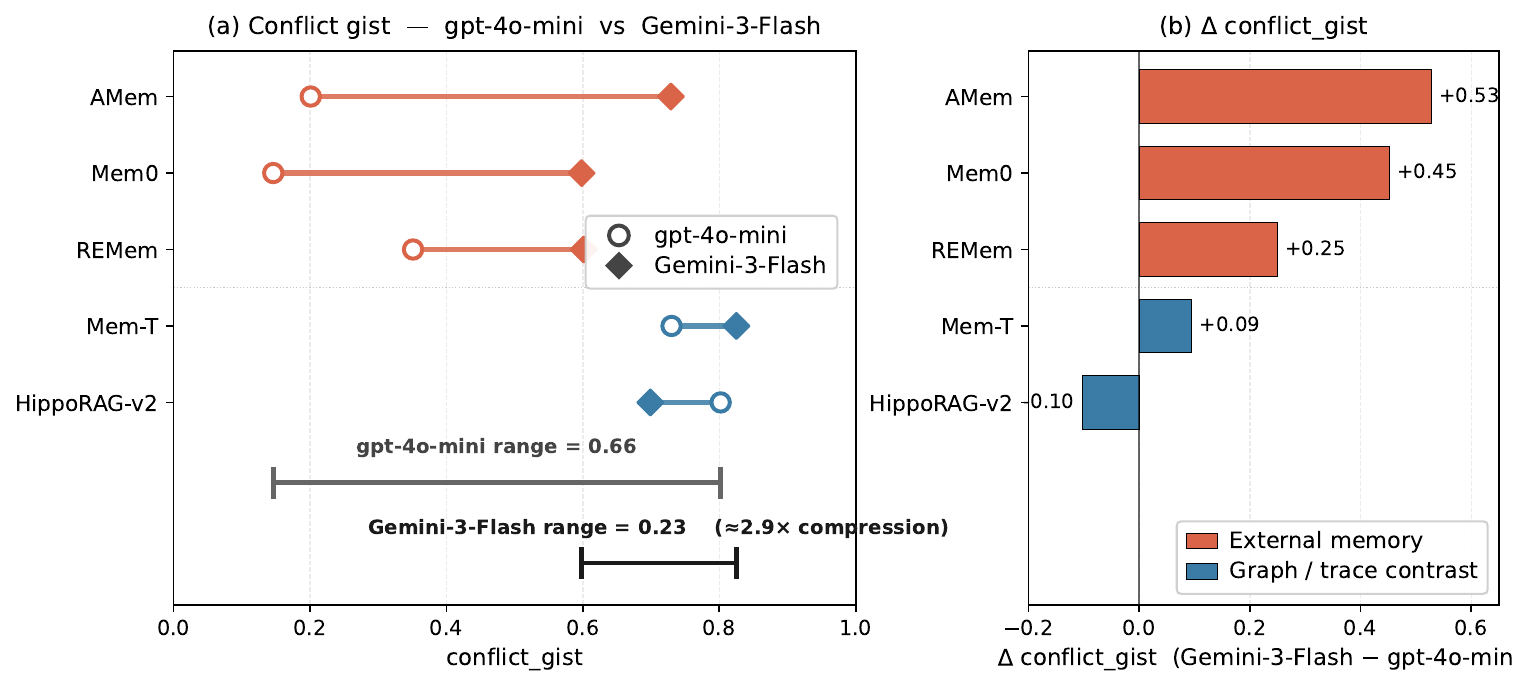}
\caption{Backbone-sensitivity panel on conflict probes. The main-adapter
gpt-4o-mini condition is the audit baseline, while the answer-time generator is
swapped to Gemini-3-Flash. }
\label{fig:gemini-5system-panel}
\end{figure*}

\begin{table*}[htbp]
\centering
\setlength{\tabcolsep}{10pt}
\resizebox{\columnwidth}{!}{%
\begin{tabular}{@{}llrrr@{}}
\toprule
System & Group & 4o-mini & Gemini & $\Delta$ \\
\midrule
HippoRAG-v2 & Graph RAG & 80.2\% & 69.9\% & -10.3 \\
Mem-T & Trace & 73.0\% & 82.5\% & +9.5 \\
REMem & External & 35.1\% & 60.1\% & +25.0 \\
AMem & External & 20.1\% & 72.9\% & +52.8 \\
Mem0 & External & 14.6\% & 59.8\% & +45.2 \\
\bottomrule
\end{tabular}}

\vspace{2pt}
\resizebox{\columnwidth}{!}{%
\begin{tabular}{@{}llrrr@{}}
\toprule
Generator & Subset & Base & Oracle & Lift \\
\midrule
4.1-mini & All-fail traj. & 0.0\% & 70.0\% & +70.0 \\
4.1-mini & Dynamic traj. & 33.8\% & 72.5\% & +38.7 \\
4.1-mini & Hard cur./hist. & 8.5\% & 75.0\% & +66.5 \\
4o-mini & All-fail traj. & 0.0\% & 85.0\% & +85.0 \\
4o-mini & Dynamic traj. & 33.8\% & 80.0\% & +46.2 \\
4o-mini & Hard cur./hist. & 8.5\% & 80.0\% & +71.5 \\
\bottomrule
\end{tabular}}
\caption{Backbone-sensitivity numeric checks. Left: Conflict Gist shifts when
the answer generator is swapped from gpt-4o-mini to Gemini-3-Flash; AMem and
Mem-T are exploratory 3-user subsets, while the other rows are full 20-user
runs. Right: oracle-evidence recoverability on the same 120-probe balanced
sample, showing large lifts under both oracle backbones. Deltas and lifts are
percentage points.}
\label{tab:gemini-conflict-shift}
\label{tab:appendix-backbone-sensitivity}
\end{table*}

\begin{table*}[t]
\centering
\resizebox{\columnwidth}{!}{%
\begin{tabular}{lll}
\toprule
Audit signal & Mem0 & HippoRAG-v2 \\
\midrule
Evidence form & Atomic facts & Raw passages \\
Returned items & 10.0 avg. facts & Variable passages \\
High-relevance fragment in top 3 & 32/40 & n/a \\
Answer begins with refutation & n/a & 32/40 \\
\bottomrule
\end{tabular}}
\caption{Retrieval-interface audit on 40 user-0 conflict probes under the
audited system pipelines. The audit illustrates differences in retrieved
evidence form and should be read as an interface check, not as a substitute for
the main rows.}
\label{tab:appendix-path2-audit}
\end{table*}

\begin{table*}[t]
\centering
\resizebox{\columnwidth}{!}{%
\begin{tabular}{lrrr}
\toprule
Mem0 setting & gpt-4o-mini & Gemini-3-Flash & Gap \\
\midrule
Vanilla Mem0 & 13.3\% & 80.4\% & +67.1 pp \\
Mem0 + score visibility & 26.3\% & 72.1\% & +45.8 pp \\
\midrule
Gap reduction & \multicolumn{3}{c}{21.3 pp, about 32\%} \\
\bottomrule
\end{tabular}}
\caption{Mem0 score-visibility intervention on paired users \{0,1,2\}. The
vanilla gpt-4o-mini value is a paired-subset baseline, not the full 20-user
Mem0 main-row value. Surfacing retrieval similarity scores reduces Mem0's
backbone-coupling on conflict Gist, with side costs in the same audit:
Gemini-3-Flash conflict hallucination rises from 7.1\% to 12.1\% and boundary
abstention falls from 88.7\% to 68.8\%.}
\label{tab:appendix-path3-score}
\end{table*}

\end{document}